\documentclass[10pt,letterpaper]{article}
\usepackage[left=1.5in,right=1.5in,top=1in,bottom=1in]{geometry}
\usepackage{times}
\usepackage[T1]{fontenc}
\usepackage[utf8]{inputenc}
\usepackage{amsmath,amsfonts,bm}

\def\eqref#1{equation~\ref{#1}}
\def\Eqref#1{Equation~\ref{#1}}
\def\1{\bm{1}}

\DeclareMathAlphabet{\mathsfit}{\encodingdefault}{\sfdefault}{m}{sl}
\SetMathAlphabet{\mathsfit}{bold}{\encodingdefault}{\sfdefault}{bx}{n}

\usepackage{amsmath,amssymb,amsthm,booktabs,array,tabularx,multirow}
\usepackage{graphicx,microtype}
\usepackage[authoryear,round]{natbib}
\setcitestyle{authoryear,round,citesep={;},aysep={,},yysep={;}}
\usepackage{titlesec}
\usepackage{placeins}
\usepackage{hyperref,url}
\titleformat{\section}{\large\bfseries}{\thesection}{0.75em}{}
\titleformat{\subsection}{\normalsize\bfseries}{\thesubsection}{0.75em}{}
\titleformat{\subsubsection}{\normalsize\bfseries}{\thesubsubsection}{0.75em}{}
\titlespacing*{\section}{0pt}{2.5ex plus 0.5ex minus 0.2ex}{1.2ex}
\titlespacing*{\subsection}{0pt}{2ex plus 0.4ex minus 0.2ex}{0.8ex}
\titlespacing*{\subsubsection}{0pt}{1.5ex plus 0.3ex minus 0.2ex}{0.6ex}
\date{}
\makeatletter
\renewcommand{\@maketitle}{%
  \begin{center}
    {\LARGE\bfseries\@title\par}
    \vspace{14pt}
    {\normalfont\normalsize\@author\par}
  \end{center}
  \vspace{8pt}
}
\makeatother

\newtheorem{proposition}{Proposition}

\graphicspath{{figures/}}
\hypersetup{
  hidelinks,
  pdftitle={The Geometry of Conceptors in Neural-Activation Spaces},
  pdfauthor={Ilias Triantafyllopoulos, Young Min Cho, Ren Tao, Miranda Muqing Miao, Sunny Rai, Lyle Ungar, Sharath Chandra Guntuku, Sunghye Cho, Tyler Moore, Neville Ryant, Jo{\~a}o Sedoc}
}

\title{The Geometry Of Conceptors In Neural-Activation Spaces}

\newcommand{\paperauthorblock}{%
  \parbox[t]{\dimexpr\textwidth-2\tabcolsep\relax}{%
    \centering
    \normalfont\normalsize
    {\bfseries
    \mbox{Ilias Triantafyllopoulos\textsuperscript{1}}\quad
    \mbox{Young Min Cho\textsuperscript{2}}\quad
    \mbox{Ren Tao\textsuperscript{2}}\\[3pt]
    \mbox{Miranda Muqing Miao\textsuperscript{2}}\quad
    \mbox{Sunny Rai\textsuperscript{2}}\quad
    \mbox{Lyle Ungar\textsuperscript{2}}\quad
    \mbox{Sharath Chandra Guntuku\textsuperscript{2}}\\[3pt]
    \mbox{Sunghye Cho\textsuperscript{4}}\quad
    \mbox{Tyler Moore\textsuperscript{2}}\quad
    \mbox{Neville Ryant\textsuperscript{3}}\quad
    \mbox{Jo{\~a}o Sedoc\textsuperscript{1}}
    \par}
    \vspace{7pt}
    {\small
    \textsuperscript{1}New York University\quad
    \textsuperscript{2}University of Pennsylvania\\[2pt]
    \textsuperscript{3}Linguistic Data Consortium,
    University of Pennsylvania\\[2pt]
    \textsuperscript{4}University of Cincinnati    \par}
  }%
}

\author{\paperauthorblock}

\begin{document}

\maketitle

\begin{abstract}
Conceptors represent concepts as graded collections of directions in neural-activation space rather than as single vectors. We develop a geometric account of how these representations depend on the contrast used to define a concept: opposing values in bipolar fitting, or presence and absence in unipolar fitting. 
%Conceptors represent activation patterns as soft linear filters that retain multiple directions with different strengths. Their geometry depends on how the anchoring examples define a concept: bipolar comparisons contrast opposing values, while unipolar comparisons contrast presence with absence. We study how this distinction shapes what conceptors preserve and amplify. 
We derive an exact decomposition of a centered pooled conceptor into an average pole filter, a mean-separation update, and a covariance-shape update. Experiments across five language models on sentiment, political leaning and stance, and depression expression show that both updates improve reconstruction of held-out contrasts. Additional directions preserve variation beyond the mean-difference line, and changing the comparator changes the benefit of these directions. 
%This geometry also has consequences for amplification: in experiments, rank-20 conceptors fit to opposing poles increase average evidence for both poles under a fixed polarity readout more consistently than presence--absence fits. The strength of the dimensions retained by the conceptor at a later activation also correlates positively with layerwise probe performance on six public benchmarks, providing a descriptive diagnostic for investigating where to probe. Cross-concept overlap quantifies geometric interference under hard subspace removal, connecting shared directions to Boolean composition. 
Shape, size, and overlap each show up downstream: in amplification experiments, bipolar fits raise evidence for both poles more consistently than unipolar fits; conceptor size (the quota) correlates positively with layerwise probe performance; and cross-concept overlap quantifies interference under subspace removal, linking shared directions to Boolean composition.
{In an additional generation evaluation, conceptor filters achieve higher target-expression win rates against an unsteered reference than three mean-based controls.} Together, these findings explain how fitting choices shape conceptor geometry and support bipolar fitting when the objective is to amplify both opposing poles.

\end{abstract}

\section{Introduction}

Conceptors represent patterns of neural activity as soft linear filters
\citep{jaeger2014,jaeger2017}, retaining learned directions with different
strengths, or gains. 
They have been used to refine language
embeddings, mitigate bias, and steer language models
\citep{liu2018correcting,karve2019conceptor,postmus2024,
triantafyllopoulos2026conceptors}. Their Boolean operations allow
represented patterns to be combined or excluded \citep{jaeger2014}.

\begin{figure}[t]
    \centering
    \includegraphics[width=\textwidth]{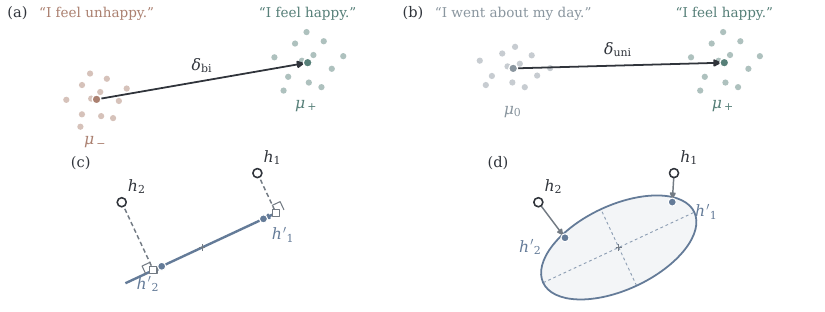}
    \vspace{-0.6cm}
    \caption{\textbf{Comparison and rank are separate choices.} Mean differences contrast (a) opposing poles or (b) presence and absence. Conceptors retain (c) one axis or (d) several, with shrinkage. Open circles: common inputs; squares: orthogonal projections; filled circles: outputs. The segment and ellipse are images of the unit ball. Coordinates and filters are illustrative.}
    \label{fig:four}
\end{figure}

A label such as ``positive sentiment'' leaves an important choice open.
Comparing ``This film is wonderful'' with ``This film is awful''
contrasts opposing evaluations. Comparing it with ``This film lasts
two hours'' contrasts an evaluation with its absence. We call these
comparisons \emph{bipolar} and \emph{unipolar}. This operational distinction is related to the distinction between naturally contrasting binary concepts and features without natural contrasts studied by~\cite{park2024}. These terms describe the fitting comparisons,
without assumptions about psychological dimensionality or independence
\citep{watson1985toward,green1993measurement}.
Figure~\ref{fig:four}(a,b) illustrates their mean-difference directions.
For conceptors, changing the comparator changes the activation
statistics used for fitting. Positive-only fitting is a separate
construction using a single collection.

The comparison and the form of the representation are separate
choices. Mean differences, commonly used in additive steering
\citep{turner2023actadd,rimsky2024caa}, summarize separation between
two averages. Conceptors also reflect variation among individual
activations, retaining one or several directions with different
gains (Figure~\ref{fig:four}(c,d)). Filtering and addition are also
distinct: filtering scales components already present in an input,
whereas fixed addition applies the same displacement to every input.

For centered pooled fits, the covariance contains both within-pole
variation and separation between pole means. We characterize how
these contributions appear in the conceptor itself and how pooling
differs from averaging separately fitted pole filters. This connects
the mean-difference line to a richer representation without assuming
that every additional direction is a distinct semantic feature:
retained variation may also reflect topic or wording.
We also examine retention strength and shared directions.
\emph{Quota}, the average conceptor gain, summarizes overall retention
strength; we study its descriptive association with layerwise probe
performance. Subspace overlap addresses a complementary question:
whether removing directions associated with one concept also removes
components retained by another, with implications for composition.

We investigate sentiment, political leaning and stance, and depression
expression across five models: Gemma-2-2B-IT, Gemma-2-9B-IT, Qwen~2.5,
Llama-3.1-8B-Instruct, and OLMo-3-7B-Instruct. We select these concepts a priori to span a spectrum from relatively simple to multifaceted constructs.
  Labeled examples for each let us test whether the same geometric
  patterns hold across different kinds of concepts.

Our contributions are:

\begin{enumerate}
    \item \textbf{Explain how conceptors represent richer geometry than mean difference.}
    We derive an exact breakdown of what changes when a conceptor is
    fitted to both poles of a concept. The separation between their averages and
    differences in how their examples vary both help preserve contrasts
    in new contexts. Additional directions capture variation that a
    single mean difference misses. {(Sections~\ref{sec:pooling} and~\ref{sec:c1:pooling}--\ref{sec:c1:state}.)}

    \item \textbf{Show why opposing examples matter for amplification.}
    Conceptors fitted to both opposing poles strengthen average
    classifier evidence for each pole more consistently than those
    fitted to presence and absence. {(Sections~\ref{sec:c1:amplification}--\ref{sec:behavioral}.)}

    \item \textbf{Connect conceptor strength to where information is most linearly observable.}
    Quota summarizes how strongly a conceptor retains activation
    patterns. Across six public benchmarks, higher quota tends to
    accompany better classifier performance across layers. {(Section~\ref{sec:c2:quota}.)}

    \item \textbf{Explain overlaps between concepts geometries.}
    Across five models, sentiment and depression-expression
    representations overlap more than either overlaps with political
    leaning. We show how overlap determines the average loss when
    one representation's directions are removed from another.
    This helps interpret the geometry behind Boolean composition. {(Section~\ref{sec:c3:overlap}.)}
\end{enumerate}

Our focus is the geometry of conceptors used for steering. {We analyze activations and frozen readouts, supplemented by a
generated-text evaluation of conceptor filtering
(Section~\ref{sec:behavioral}).}

\section{Related Work}
\label{sec:related}

\textbf{Conceptors and activation steering.}
Conceptors represent activation patterns as regularized linear filters
with Boolean operations \citep{jaeger2014,jaeger2017}.
\citet{liu2019continual} use these operations to update sentence
representations across corpora while retaining performance on earlier
corpora. For contextualized-embedding debiasing,
\citet{sedoc2019role} combine conceptors learned from homogeneous
word-list subsets through OR. \citet{yifei2023conceptor} extend
conceptor debiasing to BERT and GPT, using NOT for bias removal and AND
for intersectional bias, and introduce Conceptor-Intervened BERT.
\citet{postmus2024} apply conceptors to language-model steering and
combine steering goals through Boolean operations.
Additive methods such as ActAdd and CAA summarize a contrast with a
single displacement vector
\citep{turner2023actadd,rimsky2024caa}.

\textbf{Geometry for probing and intervention.}
\citet{park2024} connect linear concept representations,
measurement, and intervention through a causal inner product.
Their categorical extension distinguishes feature presence from
contrasts between attribute values \citep{park2025categorical}.
\citet{park2026softmax} derive probe-guided dual steering from softmax
information geometry, with regularized local updates determined by
output-unembedding covariance.
\citet{gurnee2026workspace} use averaged downstream Jacobians to
identify verbalizable representations involved in internal reasoning.
These approaches connect representation geometry to how the model
reads or uses an activation.

\textbf{Empirical concept representations.}
\citet{sofroniew2026emotions} identify emotion directions in Claude
Sonnet~4.5 and show that interventions along them influence behavior.
For depression assessment, \citet{ravenda2026navigating} use activation
steering to improve symptom-relevance prediction and questionnaire
completion. \citet{zhu2026interpretable} construct symptom directions
whose projection scores on held-out text align with clinician
annotations.
\citet{abootorabi2026steeringgeometry} find that methods with comparable
steering performance can differ substantially in how well their
geometry reflects relationships among human values. This motivates
examining both the effects of interventions and the structure of the
representations they use.

Our focus is the geometry learned by conceptors from contrasting
activation collections: how pooling changes the filter, what its
spectrum reveals, and which directions it shares with other concepts.

\section{Background}
\label{sec:bg:background}

We consider hidden activations $h\in\mathbb{R}^{d}$ at a fixed model
layer and extraction site, omitting these indices for clarity. All
geometric quantities use the Euclidean inner product in this activation
space.

\subsection{Semantic comparisons and mean differences}
\label{sec:bg:comparisons}

\begingroup
An \emph{attribute} is the semantic property under study, such as
sentiment. A \emph{pole} is one of its contrasting expression categories,
such as positive or negative sentiment. Let
$\mathcal{H}_{s}=\{h_i^{(s)}\}_{i=1}^{n_s}$ denote the activation
collection for category $s\in\{+,-,0\}$: the target pole, its opposite,
or a specified absence reference, respectively.
\endgroup
Write
$\mu_s=n_s^{-1}\sum_{i=1}^{n_s}h_i^{(s)}$ for the mean of collection
$s\in\{+,-,0\}$. The corresponding difference-of-means vectors are
\begin{equation}
    \delta_{\mathrm{bi}}=\mu_{+}-\mu_{-},
    \qquad
    \delta_{\mathrm{uni}}=\mu_{+}-\mu_{0}.
    \label{eq:bg:diffmean}
\end{equation}
The bipolar comparison contrasts opposing attribute values. The unipolar
comparison contrasts presence with absence, following the
feature--contrast distinction of \citet{park2024}.
For sentiment, these compare positive expression with negative expression
and sentiment-free expression, respectively
(Figure~\ref{fig:four}(a,b)). The absence reference is defined by an
empirical collection, whose mean $\mu_0$ can be nonzero.
For equally weighted matched pairs, each vector also equals the average
of the paired activation differences. Such contrasts underlie additive
steering \citep{turner2023actadd,rimsky2024caa}, which applies the
same displacement to each activation:
$h'=h+\beta\delta$, with strength $\beta\in\mathbb{R}$.
The vector summarizes the separation between the group means.

\subsection{Conceptors}
\label{sec:bg:estimation}

A conceptor is a linear filter learned by balancing reconstruction of
its fitting activations against regularization of the operator
\citep{jaeger2014,jaeger2017}. Conceptors have also been used for
language-model steering \citep{postmus2024}.
{Let $\mathcal{H}=\{h_i\}_{i=1}^{n}$ be the fitting activation
collection. Given a reference $c\in\mathbb{R}^{d}$, define $x_i=h_i-c$
and the second moment}
\begin{equation}
    R_c=\frac{1}{n}\sum_{i=1}^{n}x_ix_i^\top.
    \label{eq:bg:moment}
\end{equation}
The reference $c$ sets the origin for fitting. Choosing $c$ as the fitting
mean gives a centered covariance. Choosing $c=0$ gives an uncentered
second moment. We use normalization $1/n$ in both cases.

The conceptor is the solution to
\begin{equation}
    C_\lambda(R_c)
    =\underset{M\in\mathbb{R}^{d\times d}}{\arg\min}
    \left\{
        \frac{1}{n}\sum_{i=1}^{n}\|x_i-Mx_i\|_2^2
        +\lambda\|M\|_F^2
    \right\},
    \qquad \lambda>0.
    \label{eq:bg:objective}
\end{equation}
The first term encourages reconstruction of the fitting activations.
The second penalizes the size of the filter. The unique minimizer is
\begin{equation}
    C_\lambda(R_c)=R_c(R_c+\lambda I)^{-1}.
    \label{eq:bg:conceptor}
\end{equation}
The usual \emph{aperture} parameter $\alpha>0$ satisfies
$\lambda=\alpha^{-2}$.
Appendix~\ref{app:bg:solution} gives the standard derivation.

A bipolar pooled fit combines $\mathcal{H}_{+}$ and $\mathcal{H}_{-}$.
A presence--absence pooled fit combines $\mathcal{H}_{+}$ and
$\mathcal{H}_{0}$. Fitting $\mathcal{H}_{+}$ alone gives a
\emph{positive-only} conceptor. Each construction also requires a choice
of fitting reference $c$. Together, the fitting collection and reference
determine $R_c$.

\subsection{Spectral geometry}
  \label{sec:bg:geometry}

  Let $R_c=U\operatorname{diag}(r_1,\ldots,r_d)U^\top$, with
  $r_1\geq\cdots\geq r_d\geq0$ and orthonormal eigenvectors $u_j$.
  Then
  \begin{equation}
      C_\lambda(R_c)=\sum_{j=1}^{d}g_j u_ju_j^\top,
      \qquad
      g_j=\frac{r_j}{r_j+\lambda}\in[0,1).
      \label{eq:bg:gains}
  \end{equation}
  At fixed $R_c$, increasing aperture raises positive gains without
  changing the axes or exact rank.

  The leading-$k$ approximation
  $C_{\lambda,k}=\sum_{j=1}^{k}g_j u_ju_j^\top$ filters via
  $h'=c+C_{\lambda,k}(h-c)$, projecting $h-c$ onto the retained axes
  and shrinking its coordinates (Figure~\ref{fig:four}(c,d)).
  The leading axis need not align with the mean difference
  (Appendix~\ref{app:pool:rankone}).
  Additive steering applies the same displacement $\beta\delta$
  to every input.

  The \emph{quota} averages gains over all $d$ activation dimensions
  \citep{jaeger2014}:
  \begin{equation}
      q(C)=\frac{\operatorname{tr}(C)}{d}
          =\frac{1}{d}\sum_{j=1}^{d}g_j,
      \qquad C=C_\lambda(R_c).
      \label{eq:bg:quota}
  \end{equation}
  Quota depends on the fitted moment and aperture; its association
  with probe performance is empirical. 
  For hard orthogonal projectors, conceptor NOT, AND, and OR give
  the orthogonal complement, intersection, and combined span
  \citep{jaeger2014}; soft operations also depend on gains.
  Appendix~\ref{app:bg:overlap} relates overlap to average squared
  length retained under hard removal.
\section{How pooling shapes a conceptor}
\label{sec:pooling}

Pooling both poles combines their separation with variation within each
pole. We characterize these contributions to the filter and its
reduction to the mean-difference line.

Let $R_+,R_-$ be the covariances of the two poles, each centered on its
own mean, and let $p,q>0$ be their weights, with $p+q=1$.
Write $\delta=\mu_+-\mu_-$ and $\mu=p\mu_++q\mu_-$.
Centering the pooled examples on $\mu$ gives
\begin{equation}
    R_{\mathrm{pool}}
    =\underbrace{pR_++qR_-}_{R_{\mathrm{w}}}
     +pq\delta\delta^\top.
    \label{eq:pool:covariance}
\end{equation}
The terms describe within-pole spread and mean separation.
Empirical covariances use $1/n_s$ normalization; ordinary pooling sets
$p=n_+/(n_++n_-)$ and $q=1-p$.
{Here $p$ and $q$ specify the relative weight assigned to each
pole. Our grouped experiments instead weight source groups equally and
the two poles equally within each group
(Appendix~\ref{app:c1:protocol}).}
All filters below use the same numerical $\lambda>0$.

\begin{proposition}[Pooling decomposition]
\label{prop:pool:decomposition}
Let $A=R_{\mathrm{w}}+\lambda I$ and
$\bar C=pC_\lambda(R_+)+qC_\lambda(R_-)$.
The full pooled conceptor satisfies
\begin{equation}
    C_\lambda(R_{\mathrm{pool}})
    =\underbrace{\bar C}_{\text{average pole filter}}
     +\underbrace{D_\mu}_{\text{mean separation}}
     +\underbrace{J_R}_{\text{covariance shape}},
    \label{eq:pool:decomposition}
\end{equation}
where
\begin{align}
    D_\mu
    &=\frac{\lambda pq}
    {1+pq\delta^\top A^{-1}\delta}
    (A^{-1}\delta)(A^{-1}\delta)^\top,
    J_R=C_\lambda(R_{\mathrm{w}})-\bar C.
    \label{eq:pool:shape-update}
\end{align}
Both updates are positive semidefinite. The mean update has rank one
when $\delta\ne0$, and
$\operatorname{rank}(J_R)=\operatorname{rank}(R_+-R_-)$.
\end{proposition}

\textbf{The average pole filter.}
The weighted average $\bar C$ combines filters fitted after subtracting
each pole's own mean. It represents within-pole variation and can
already retain several directions.

\textbf{The mean-separation update.}
For distinct means, $D_\mu$ is a rank-one update along $A^{-1}\delta$~\citep{fisher1936use}.
This adjusts the mean difference by downweighting components along
principal directions with greater within-pole spread. It follows the
mean-difference line when $\delta$ is an eigenvector of
$R_{\mathrm{w}}$.

\textbf{The covariance-shape update.}
Fitting the average covariance differs from averaging the fitted
filters because the conceptor depends nonlinearly on covariance.
Thus $J_R$ reflects differences in the poles' spread and orientation. It vanishes
exactly when $R_+=R_-$ and can involve several directions even when
the means coincide.

The identity applies to full filters before rank truncation, at a shared
$\lambda$, including presence--absence comparisons.
Appendix~\ref{app:pool:decomposition} gives the proof.
Fitting only two equally weighted, distinct pole means about their
  midpoint gives a rank-one conceptor along the mean difference;
  within-pole variation outside this line increases the rank of the
  full pooled conceptor (Appendix~\ref{app:pool:rankone}).

\section{Data and measurements}
\label{sec:data}

\subsection{Datasets}
\label{sec:data:datasets}

\textbf{BipolarConcepts.}
Our original fitting collection contrasts positive/negative sentiment,
left/right political leaning, and depression expression/counter-expressions.
The datasets contain 100 pairs for each concept (600 sentence entries). These supply the original
conceptor geometry; Appendix~\ref{app:data:authored} includes more details.

\textbf{ConceptVariants.}
This authored collection contains 312 sentences covering 24 topics per
concept (sentiment, depression expression, and political stance). Each sentiment or policy topic has two opposing and two neutral
formulations. Each symptom topic has one symptom formulation and four
comparators. We vary the reference within a topic to examine how the
comparison shapes the representation. The available sentence inventories
have no exact or normalized-text overlap with \textsc{BipolarConcepts};
these are separate collections, not nested subsets.

\textbf{ConceptGroups.}
This grouped evaluation suite combines 1,400 reviews from three domains (sentiment concept)
\citep{kotzias2015sentiment}, 196 authored sentences across fourteen
symptom facets (depression concept), and 994 favor/against tweets across five stance topics (political stance concept)
\citep{mohammad2016stance,barbieri2020tweeteval}.
Each group is balanced across its two labels. We fit representations on
the remaining groups and evaluate the excluded group's mean difference.

\textbf{Public probing benchmarks.}
We train and evaluate probes on SST-2 \citep{socher2013sst}, Rotten
Tomatoes \citep{pang2005rotten}, TweetEval sentiment
\citep{barbieri2020tweeteval}, Hyperpartisan News
\citep{kiesel2019hyperpartisan}, Reddit Depression
\citep{pixel42depression}, and Cross-cultural Reddit
\citep{rai2025crosscultural}. Their probe scores are compared with
conceptor spectra fitted on \textsc{BipolarConcepts}.

Appendix~\ref{app:data} gives the collection inventories, comparison
labels, evaluation partitions, and available length statistics.

\subsection{Measurements}
\label{sec:data:measurements}

For a nonzero mean difference $\delta$, an orthonormal basis
  $U_k\in\mathbb{R}^{d\times k}$, and a filter $M$, define
  \begin{align}
      \operatorname{Coverage}_k(\delta)
      =\frac{\|U_k^\top\delta\|_2^2}{\|\delta\|_2^2}, E(M,\delta)
      =\frac{\|M\delta-\delta\|_2^2}{\|\delta\|_2^2}.
      \label{eq:data:error}
  \end{align}
  \emph{Coverage} measures the fraction of squared length retained by
  projection; \emph{reconstruction error} also reflects filter gains.
  Fidelity $F=1-E$ equals coverage for $M=U_kU_k^\top$.

  For orthonormal rank-$k$ bases $U_A,U_B$, define
  \begin{equation}
      \operatorname{Overlap}(A,B)
      =\frac{1}{k}\|U_A^\top U_B\|_F^2.
      \label{eq:data:overlap}
  \end{equation}
  Overlap is zero for orthogonal subspaces and one for identical subspaces.

  Spectral summaries are quota (\Eqref{eq:bg:quota}) and
  $\operatorname{EVR}_k=\sum_{j=1}^{k}r_j/\sum_{j=1}^{d}r_j$,
  the leading-$k$ share of the eigenvalue sum for a nonzero spectrum.
  Logistic-probe ROC-AUC measures label separability; layerwise Pearson
  correlations relate it to these summaries.
  Each experiment specifies rank, fitting reference, regularization,
  and evaluation split.

\section{Experiments}
\label{sec:experiments}

We use Gemma-2-2B-IT, Gemma-2-9B-IT, Qwen-2.5-3B-Instruct, Llama-3.1-8B-Instruct, and
OLMo-3-7B-Instruct.
The default features are mean-token activations at midpoint blocks
13, 21, 18, 16, and 16. The main multi-direction comparisons use $k=20$, with $k=1$ as a
  single-direction control; the primary decomposition ablation and quota
  use the full spectrum.
\textsc{ConceptGroups} evaluations exclude one complete group from
fitting and average held groups equally. Appendix~\ref{app:c1} gives
protocols and controls. {Midpoint is a fixed comparison site, not a selected
optimum. All-layer and final-block controls test the dependence on depth
(Appendix~\ref{app:c1:layers}); final-layer coverage advantages over the
mean line range from 14.5 to 56.2 percentage points.}

\subsection{Pooling and the fitting reference}
\label{sec:c1:pooling}

We compare pooled and separate-pole rank-20 supports on
\textsc{ConceptGroups} to test the effect of pooling at equal fitting
counts. Each fit receives the same number of source sentences, and we
evaluate coverage of the excluded group's mean difference. We center
on either each fit's own mean or a shared source center.
Table~\ref{tab:c1:centering} shows higher pooled coverage than the
positive-only fit in all fifteen own-centered model/data conditions
and than the negative-only fit in fourteen. The exception is OLMo
Pol.~Stance, where pooled and negative-only coverage are 30.33\% and
30.61\%. With a shared source center, pooling wins in four conditions
against the positive pole and six against the negative pole.
Retaining each pole's mean offset therefore substantially reduces the
pooling advantage. Appendix~\ref{app:c1:pooling} gives the original
fitting-collection curves, where pooled leading-ten capture has the
highest layer average in all fifteen model--concept pairs. On \textsc{BipolarConcepts}, pooled rank-20 support has the highest
  layer-average coverage in all fifteen model--concept pairs, exceeding
  one half in each and reaching 97.4--99.7\% for Sentiment
  (Appendix~\ref{app:c1:pooling}).
\begingroup
A topic-matched control further tests what these supports preserve.
For sentiment and political leaning, concept-expressing supports exceed
both equal-count neutral controls in all 40 model--concept--site--pooling
settings. This supports target-related structure beyond shared topic
content
(Appendix~\ref{app:controls:neutral}).
\endgroup

\begin{table}[t]
\centering
\footnotesize
\setlength{\tabcolsep}{3pt}
\begin{tabular}{@{}llrrrrrr@{}}
\toprule
& & \multicolumn{3}{c}{Own mean} & \multicolumn{3}{c}{Shared source center} \\
\cmidrule(lr){3-5}\cmidrule(l){6-8}
Model & Data & Pool & $+$ & $-$ & Pool & $+$ & $-$ \\
\midrule
Gemma 2 2B & Sentiment & \textbf{75.5} & 52.3 & 48.4 & 75.4 & 78.1 & \textbf{80.2} \\
 & Pol. Stance & \textbf{51.7} & 49.8 & 50.8 & \textbf{51.7} & 50.1 & 51.0 \\
 & Depression & \textbf{69.8} & 51.1 & 52.7 & 69.8 & \textbf{70.6} & 69.6 \\
\midrule
Gemma 2 9B & Sentiment & \textbf{74.3} & 54.4 & 52.6 & 74.3 & 76.3 & \textbf{77.9} \\
 & Pol. Stance & \textbf{48.5} & 47.7 & 47.8 & \textbf{48.5} & 48.3 & 48.2 \\
 & Depression & \textbf{73.0} & 57.5 & 57.5 & 73.1 & \textbf{73.9} & 73.0 \\
\midrule
Qwen 2.5 3B & Sentiment & \textbf{96.0} & 94.0 & 93.8 & 96.0 & 96.5 & \textbf{96.7} \\
 & Pol. Stance & \textbf{74.0} & 73.8 & 72.6 & 73.9 & \textbf{74.0} & 72.7 \\
 & Depression & \textbf{92.2} & 88.5 & 89.4 & 92.2 & \textbf{92.5} & 92.4 \\
\midrule
Llama 3.1 8B & Sentiment & \textbf{84.6} & 76.8 & 75.8 & 84.6 & 85.5 & \textbf{86.3} \\
 & Pol. Stance & \textbf{57.7} & 56.8 & 56.9 & \textbf{57.6} & 57.4 & 57.3 \\
 & Depression & \textbf{81.6} & 74.2 & 74.4 & 81.7 & \textbf{82.6} & 81.9 \\
\midrule
OLMo 3 7B & Sentiment & \textbf{42.7} & 21.8 & 11.4 & 42.7 & 48.3 & \textbf{49.6} \\
 & Pol. Stance & 30.3 & 29.9 & \textbf{30.6} & 30.4 & 30.3 & \textbf{30.9} \\
 & Depression & \textbf{30.3} & 11.8 & 14.2 & 30.3 & \textbf{32.0} & 31.2 \\
\bottomrule
\end{tabular}
\caption{Held-group rank-20 coverage (\%) at equal fitting counts.
  Pool fits both poles; $+$/$-$ fit the positive/negative pole separately.
  Bold marks the highest coverage within each model--concept row and
  centering condition; higher coverage indicates better preservation
  of the held-group contrast.}
\label{tab:c1:centering}
\end{table}

% increase the caption with conclusion. What bold means in this table  

\subsection{Testing the two pooling updates}
\label{sec:c1:ablation}

\begin{figure}[t]
    \centering
    \includegraphics[width=0.96\linewidth]
        {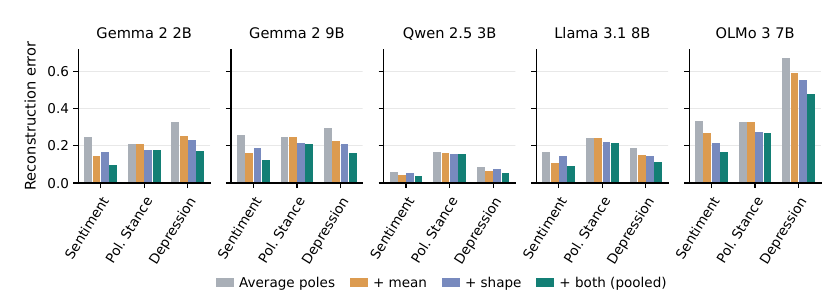}
    \caption{\textbf{Reconstruction under shared regularization.}
     Midpoint \textsc{ConceptGroups} error, averaged equally over held groups;
     lower is better.}
    \label{fig:c1:ablation}
\end{figure}

We remove the decomposition terms to test their contributions to
held-contrast reconstruction. The operators $\bar C$,
$\bar C+D_\mu$, $\bar C+J_R$, and $C_{\mathrm{pool}}$ share the
source examples and $\lambda=\operatorname{tr}(R_{\mathrm{pool}})/d$.
Each receives the same held-group mean difference.
Figure~\ref{fig:c1:ablation} shows that combining both updates gives
the lowest reconstruction error in every model/data condition.
Across both sites and all three shrinkage scales, the full filter
outperforms every ablation in all 270 comparisons
(Appendix~\ref{app:c1:ablation}). At the primary midpoint setting,
the mean update contributes more for Sentiment in four of five models;
OLMo instead benefits more from the covariance-shape update. The shape
update contributes more for Pol.~Stance in all five models.
For Gemma-2B Pol. Stance, adding the shape update reduces error from
0.209 to 0.179; adding the mean update reduces it to 0.207.
These reconstruction gains are measured at shared $\lambda$. A follow-up comparison of rank-20
     orthogonal projectors fixes rank, trace, and nonzero gains: the pooled
     projector achieves higher held-contrast coverage in 251 of 270 comparisons {with a mean advantage of 10.64 percentage points}
     (Appendix~\ref{app:c1:ablation}).

 \textbf{Variation across layers.}
  At some early Qwen layers, the mean line approaches pooled rank-20
  coverage for Sentiment and Depression; the gap can be larger later.
  Both Gemma models show substantial coverage gaps throughout depth,
  while Llama and OLMo vary by task and layer
  (Appendix~\ref{app:c1:layers}).

\subsection{Retaining individual states from both poles}
\label{sec:c1:state}

We next ask how much of each sentence's activation survives the filter. {We compare $M\in\{uu^\top,C_1,C_{20}\}$, where
$u=\delta/\|\delta\|_2$ is the source mean-difference direction. $C_1$ and $C_{20}$ denote the conceptors, where only 1 or 20 leading directions are retained.}
For a held state $x=h-c$, centered on the source midpoint, we measure
$\|Mx\|_2^2/\|x\|_2^2$ and average these ratios within each pole and
held group. On \textsc{ConceptGroups}, $C_{20}$ retains
  more energy at both poles than the mean-line projector
  (Table~\ref{tab:c1:state}). $C_1$ does so in 29 out of 30
individual-pole comparisons; it falls slightly below the mean line for
OLMo's negative Sentiment pole.
Even $C_1$ almost always retains more
  energy than the mean line, despite shrinkage.  The mean line retains more positive-pole energy for Depression
and more negative-pole energy for Sentiment and Pol. Stance. Averaging over the two poles, moving from $C_1$ to $C_{20}$ increases
  retained energy by 7.2 percentage points for Sentiment, 15.1 for
  Pol.~Stance, and 13.2 for Depression.
  Additional directions therefore preserve substantially more held-state
  variation, especially for Pol.~Stance and Depression, which are by nature more complicated concepts than Sentiment.

\begin{table}[!htbp]
\centering
\footnotesize
\setlength{\tabcolsep}{3pt}
\begin{tabular}{@{}llll@{}}
\toprule
Data & Mean line & $C_1$ & $C_{20}$ \\
\midrule
Sentiment ($\uparrow$) & 35.7 / 36.4 & 61.1 / 61.1 & \textbf{68.3} / \textbf{68.4} \\
Pol. Stance ($\uparrow$) & 23.2 / 24.7 & 35.1 / 37.6 & \textbf{50.7} / \textbf{52.3} \\
Depression ($\uparrow$) & 40.8 / 36.5 & 49.7 / 44.8 & \textbf{62.1} / \textbf{58.7} \\
\bottomrule
\end{tabular}
\caption{Held-state retention (\%). Positive/negative retained energy, averaged over held groups then the five models.}
\label{tab:c1:state}
\end{table}

\subsection{Quota Correlation}
\label{sec:c2:quota}

\begin{figure}[!htbp]
    \centering
    \includegraphics[width=0.9\linewidth]{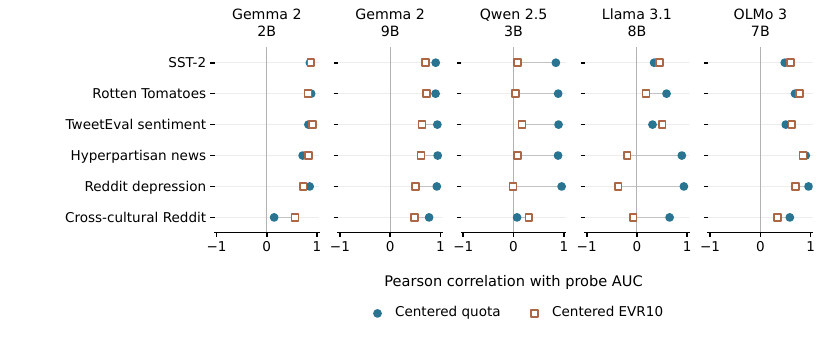}
    \vspace{-0.5cm}
    \caption{\textbf{Quota and layerwise separability.} Pearson correlations of public-benchmark probe ROC-AUC with quota and EVR$_{10}$ fitted on \textsc{BipolarConcepts}. Values closer to $+1$ indicate stronger positive tracking of probe
  AUC across layers.}
    \label{fig:c2:quota}
\end{figure}

We ask whether conceptor retention strength tracks label separability
on separate text collections. At each layer, conceptors are fitted to
the pooled poles of \textsc{BipolarConcepts} with aperture $\alpha=10$.
Separately, we train an $\ell_2$-regularized logistic probe on each
public benchmark's training split and evaluate ROC-AUC on its held-out
split. We correlate the archived layerwise AUC curves with quota and
the leading-ten explained-variance ratio
(Figure~\ref{fig:c2:quota}).
Quota correlates positively with AUC in all 30 model--benchmark
comparisons and exceeds EVR$_{10}$ in 20. 
A possible explanation for the pronounced gap in Qwen we observe is that a few
  principal components dominate its activation variance. EVR$_{10}$ can
  then remain close to one across layers, providing little contrast
  between their spectra. Quota combines gains across the full spectrum,
  allowing intermediate-strength directions to influence the summary
  even when the leading components dominate the variance.
Quota can be computed directly from the fitted conceptors without
  training another classifier. It complements information-theoretic and
  geometric diagnostics of layerwise representations \citep{skean2025layer}.
  Positive layerwise correlation alonex does not establish a
  reliable rule for selecting probe layers (Appendix~\ref{app:c2:diagnostics}, which
  also reports entropy and effective-rank controls and absolute AUC curves).

\subsection{Overlap and Boolean composition}
\label{sec:c3:overlap}

We examine which concept representations share directions and how this
affects their interaction under removal. Using \textsc{BipolarConcepts},
we compare the leading 20 conceptor axes at every recorded layer and
average overlap across layers. Sentiment and Depression expression share
more leading structure than either shares with political leaning in all
five models (Table~\ref{tab:c3:overlap}). Their layer-average overlaps are 0.318 in Llama and 0.221 in OLMo,
  compared with 0.096--0.108 and 0.035--0.042 for the respective
  political pairings. The ordering is
consistent, while the separation varies substantially across families. {The same ordering holds at all 168 recorded model--layer settings
(Appendix~\ref{app:controls:overlap}).}

\begin{table}[!htbp]
\centering
\small
\setlength{\tabcolsep}{5pt}
\begin{tabular}{@{}lrrrrr@{}}
\toprule
Concept pair & Gemma 2 2B & Gemma 2 9B & Qwen 2.5 3B & Llama 3.1 8B & OLMo 3 7B \\
\midrule
Sentiment--Depression & \textbf{0.321} & \textbf{0.300} & \textbf{0.332} & \textbf{0.318} & \textbf{0.221} \\
Sentiment--Pol. Stance & 0.154 & 0.134 & 0.169 & 0.096 & 0.035 \\
Depression--Pol. Stance & 0.163 & 0.143 & 0.182 & 0.108 & 0.042 \\
\bottomrule
\end{tabular}
\caption{Leading-twenty overlap, averaged across layers.}
\label{tab:c3:overlap}
\end{table}

Overlap (\Eqref{eq:data:overlap}) gives an exact interpretation of hard subspace removal.
Thus, the measured overlaps imply 81.8--96.5\% retained squared length
for the uniformly weighted axes of the other concept when removing
political-leaning directions. Removal between Sentiment and Depression
retains 66.8--77.9\%. These are layer-average geometric consequences
of the measured overlaps.
In the hard-projector setting, conceptor NOT uses the orthogonal
complement, AND retains intersections, and OR combines spans
\citep{jaeger2014}. The observed pair differences quantify interference
under the complement operation and provide geometric evidence relevant
to composition. 

\subsection{Bipolar vs Unipolar Capture}
\label{sec:c1:amplification}

We compare fixed mean-difference addition and centered conceptor
  amplification under bipolar and unipolar fitting, examining how
  each intervention changes evidence for the positive and opposite
  poles. Bipolar fitting uses positive and opposite examples;
  unipolar fitting pairs the positive pole with a neutral or absence
  reference. Fixed addition translates all inputs along the fitted
  mean difference, while conceptor amplification depends on each
  input's position relative to the fitting midpoint.

For each fit, let $u$ be its normalized mean difference, $c$ its
midpoint, and $C_k$ its conceptor. We compare
\begin{equation}
    T_{\mathrm{add}}(h)=h+Du,
    \qquad
    T_{\mathrm{amp}}(h)=h+tC_k(h-c),
    \label{eq:c1:actions}
\end{equation}
with $k\in\{1,20\}$, $D=\beta s$, and
  $t=D/\operatorname{RMS}_{s}(C_k(h-c))$.
  Here $s$ is a source-calibrated reference scale shared across
  methods, and $\operatorname{RMS}_{s}$ denotes weighted RMS
  over source calibration examples
  (Appendix~\ref{app:c1:amplification}).
Four grouped \textsc{ConceptVariants} folds separate fitting and
evaluation. Both fits are evaluated on the same held positive and
opposite examples using a frozen, source-trained polarity readout $f$.
Changes in standardized correct-pole scores are
\begin{equation}
\begin{aligned}
    \Delta M_+ = \mathbb{E}_+[f(T(h))-f(h)], 
    \Delta M_- = -\mathbb{E}_-[f(T(h))-f(h)].
\end{aligned}
\label{eq:c1:margin-changes}
\end{equation}
Positive values indicate stronger evidence for the corresponding pole.

For two-sided amplification, the fitting contrast matters.
  Fixed mean-difference addition favors the positive pole on average,
  whereas bipolar $C_{20}$ strengthens both mean margins in all thirty
  calibrated conditions, compared with ten for bipolar $C_1$ and seven
  for unipolar $C_{20}$ (Table~\ref{tab:c1:amplification}).
  Mean-line controls fitted against a reference do not reproduce this
  consistency, whether their presence collection contains only the
  positive pole or both poles together
  (Appendix~\ref{app:c1:amplification}).
  Simply including examples from both poles is therefore insufficient:
  how their relationship is represented matters.
  These results support bipolar conceptor fitting when the goal is to
  strengthen each pole on its respective examples.
  An input-dependent bipolar mean-line amplifier also succeeds in all
  condition averages, but $C_{20}$ achieves two-sided improvement more
  consistently across held folds (Appendix~\ref{app:c1:amplification}).

\begin{table}[!htbp]
\centering
\small
\setlength{\tabcolsep}{4pt}
\begin{tabular}{@{}lrrcrrc@{}}
\toprule
& \multicolumn{3}{c}{Bipolar fit} & \multicolumn{3}{c}{Unipolar fit} \\
\cmidrule(lr){2-4}\cmidrule(l){5-7}
Method & \shortstack{Positive\\pole} & \shortstack{Opposite\\pole} & \shortstack{Both\\improve} & \shortstack{Positive\\pole} & \shortstack{Opposite\\pole} & \shortstack{Both\\improve} \\
\midrule
Mean addition & +0.2942 & -0.2942 & 0/30 & +0.0550 & -0.0550 & 0/30 \\
$C_1$ amplification & +0.0018 & +0.0002 & 10/30 & +0.0177 & -0.0153 & 2/30 \\
$C_{20}$ amplification & +0.0301 & +0.0322 & 30/30 & +0.0177 & -0.0083 & 7/30 \\
\bottomrule
\end{tabular}
\caption{Mean standardized held-pole margin changes at $\beta=0.10$ across five models.}
\vspace{-0.3cm}
\label{tab:c1:amplification}
\end{table}

\begingroup
\subsection{Generated-text validation}
\label{sec:behavioral}
We additionally evaluate generated responses across the five models on
596 held-out inputs per model: 200 public sentiment reviews, 200 political
statements, and 196 depression-expression/wellbeing examples. This
exploratory extension fits operators to query-token activations and applies repeated interventions
during generation. Full and rank-one conceptor filters are compared with
fixed mean-difference addition, pooled-mean addition, and input-dependent
mean-line amplification. Prompting and eight-shot ICL provide text-based
reference conditions. The training details, the operators, calibration, splits, and decoding settings are specified in~Appendix~\ref{app:generation}.

A blinded GPT-6-Luna judge compares each response with its corresponding
unsteered BASE response in both presentation orders, preferring any
discernibly stronger expression of the input's target pole. The win rate
is reported, while the ties are kept as non-wins. Table~\ref{tab:behavioral_model_average}
averages models equally within each pole; Appendix~\ref{app:generation}
reports per-model results, uncertainty, and the complete rubric.

Both conceptor variants achieve higher BASE-relative win rates than all
three mean-based controls in every model--concept--pole condition.
The full and rank-one filters average 47.1\% and 50.1\%, compared with
13.4\% for mean-difference addition, 14.4\% for pooled-mean addition,
and 11.8\% for mean-line amplification. Rank-one filtering is competitive
with the full filter, while prompting and ICL reach 74.4\% and 75.1\%.
These results connect conceptor interventions to observable changes in
generated expression, complementing the geometric analysis.
\endgroup
% Load revision_colors.tex in the preamble.
\begin{table}[t]
\begingroup
\centering
\small
\setlength{\tabcolsep}{3pt}
\begin{tabular}{lrrrrrrr}
\toprule
Method & Sent.+ & Sent.$-$ & Left & Right & Dep. & Wellbeing & Mean \\
\midrule
Full conceptor filter & 55.5 & 47.2 & 40.2 & 32.5 & 48.2 & 58.8 & 47.1 \\
Rank-one conceptor filter & 58.4 & 52.5 & 41.3 & 34.8 & 52.0 & 61.4 & 50.1 \\
Fixed mean-difference addition & 16.3 & 12.9 & 13.7 & 10.7 & 13.7 & 13.3 & 13.4 \\
Pooled-mean addition & 15.0 & 15.8 & 15.4 & 11.7 & 12.9 & 15.6 & 14.4 \\
Mean-line amplification & 11.0 & 11.0 & 13.5 & 8.8 & 10.9 & 15.5 & 11.8 \\
\midrule
Eight-shot ICL & 79.8 & 86.7 & 45.8 & 63.5 & 91.3 & 83.7 & 75.1 \\
Prompt-only amplification & 83.6 & 78.0 & 75.8 & 67.9 & 90.6 & 50.2 & 74.4 \\
\bottomrule
\end{tabular}
\caption{Target-expression wins against BASE (\%). Each pole averages five models equally; Mean averages
the six poles equally. Per-model results and source-cluster intervals:
Appendix~\ref{app:generation}.}
\label{tab:behavioral_model_average}
\endgroup
\end{table}

\section{Conclusion}
  \label{sec:conclusion}

  Our exact pooling decomposition separates the average pole filter
  from mean-separation and covariance-shape contributions. Across five
  models, both contributions improve held-out contrast reconstruction
  under shared regularization. Additional directions preserve variation
  beyond the mean line, with benefits depending on model and layer.
  Reconstructing activations may require more directions than influencing
  behavior. Our findings are compatible with intervention evidence for
  a refusal-mediating direction \citep{arditi2024refusal} and
  mean-difference truth directions that can generalize across datasets
  and causally affect outputs \citep{marks2023geometry}.
  For pole prediction, features from the rank-one mean component
  $D_\mu$ outperform pooled rank-twenty features in 25 of 30
  model/data/site conditions using mean-token activations
  (Appendix~\ref{app:c1:ablation}).
  In offline amplification, bipolar rank-twenty conceptors increase both
  mean held-pole readout margins more consistently than presence--absence
  fits. Fixed addition favors one pole; input-dependent updates can
  increase both mean margins. {These offline results concern edited activations and frozen
readouts; the separate generation experiment evaluates target-pole
expression (Section~\ref{sec:behavioral})}.
  Quota's observed association with layerwise label separability
  motivates hypotheses about where to probe. Cross-concept overlap
  quantifies interference under hard subspace removal, informing
  Boolean composition. Identifying effective steering layers and
  establishing the semantic effects of Boolean combinations require
  direct intervention studies.

\subsection*{AI use statement}

We used generative AI tools to assist with drafting and revising
sections of the manuscript, improving clarity and organization, and
identifying related literature. The authors reviewed the original
sources themselves and assessed their relevance and how they were
represented in the paper.

AI-assisted coding tools supported the development and revision of
experimental code, the execution of experiments, and the implementation
of robustness and sensitivity analyses reported in the appendices.
Generative AI also assisted with refining experimental questions,
discussing the interpretation of results, and writing and explaining
mathematical derivations and proofs.

The authors reviewed the AI-assisted text, mathematical arguments,
experimental code, and reported results. This review included checking
the correspondence between the described methods and their
implementation, examining experimental outputs, and verifying
references against the original sources. Final decisions about the
methodology, interpretation, and presentation were made by the authors.
We take full responsibility for the final content of this work,
including all text, claims, code, figures, and other artifacts produced
with generative AI assistance.

\subsection*{Ethics statement}

This work examines concept representations in language-model activations
using authored sentence collections and existing public datasets.
Depression-related labels describe linguistic expressions, and the
public mental-health benchmarks use proxy labels. Our results do not
establish clinical validity for diagnosing depression, assessing suicide
risk, or inferring an individual's mental health. Throughout the paper,
``bipolar'' and ``unipolar'' describe semantic comparisons. The labels
``positive'' and ``negative'' identify the sides of a comparison and
express no judgment about their desirability.

\textbf{Content warning:} The appendix includes examples referring to
depression and suicidal ideation, which some readers may find distressing.
The authored examples document the language covered by the experimental
collections and make the comparisons transparent. They are included for
research purposes and provide no instructions for self-harm. Our use of
``amplification'' refers to {changes in model activations, readout
scores, or generated expression}; it does not advocate intensifying psychological distress or
encouraging harmful behavior.

Mental-health disclosures in public social-media datasets raise privacy
concerns even when the text is publicly accessible. Public availability
should not be interpreted as consent to sensitive profiling. Reuse and
redistribution should respect source access conditions and minimize
exposure of identifiable or sensitive text. The binary labels and
authored reference sentences also simplify diverse experiences and
viewpoints. Their interpretation may depend on language, culture, and
context, and the experiments do not establish fairness across populations.

Methods for modifying concept representations could be misused for
intrusive profiling, manipulation, or amplification of harmful content.
{Our experiments evaluate activation geometry, frozen readouts, 
  and generated target-pole expression; they do not establish deployment 
  safety.}
Applications involving users, especially in mental-health settings,
would require separate safety evaluation, appropriate oversight, and
safeguards against unintended effects.

\subsection*{Reproducibility statement}

The assumptions and theoretical results are presented in
Section~\ref{sec:pooling}, with complete proofs in
Appendix~\ref{app:math}. Appendix~\ref{app:data} documents the
sentence collections, public dataset sources, preprocessing, and
evaluation splits. Appendix~\ref{app:c1} provides experimental
settings, fitting procedures, and additional robustness analyses.
We will publicly release our codebase and the synthetic sentiment
and political-stance sentence collections, together with experiment
configurations and instructions for reproducing the analyses.
{The release will also include generation outputs and the judge
protocol, with sensitive depression-expression material subject to the
same gated-access conditions described below.}
The synthetic depression-expression collections will be distributed
through gated access because they contain sensitive mental-health
content, including references to suicidal ideation. The release
documentation will explain the access procedure and conditions of
use so that researchers can request these data for replication.



\appendix
\section{Proofs and supporting derivations}
\label{app:math}

\subsection{Derivation of the conceptor solution}
\label{app:bg:solution}

For completeness, we derive the standard conceptor solution from the
regularized reconstruction objective of \citet{jaeger2014,jaeger2017}.
Let $x_i=h_i-c$ and $R=R_c=n^{-1}\sum_i x_ix_i^\top$.
Writing $L(M)$ for the objective in \Eqref{eq:bg:objective},
we have
\begin{align}
    L(M)
    &=\frac{1}{n}\sum_{i=1}^{n}\|x_i-Mx_i\|_2^2
      +\lambda\|M\|_F^2 \nonumber\\
    &=\operatorname{tr}\!\left((I-M)R(I-M)^\top\right)
      +\lambda\operatorname{tr}(MM^\top) \nonumber\\
    &=\operatorname{tr}(R)-2\operatorname{tr}(MR)
      +\operatorname{tr}\!\left(M(R+\lambda I)M^\top\right).
    \label{eq:app:bg:expanded-objective}
\end{align}
Here we use symmetry of $R$ and the cyclic property of the trace.
Differentiating with respect to $M$ gives
\begin{equation}
    \nabla_M L=2M(R+\lambda I)-2R.
    \label{eq:app:bg:gradient}
\end{equation}
Setting this gradient to zero yields $M(R+\lambda I)=R$.
Since $R\succeq0$ and $\lambda>0$, the matrix $R+\lambda I$ is positive
definite and hence invertible. Thus the stationary point is
\begin{equation}
    C_\lambda(R)=R(R+\lambda I)^{-1}.
    \label{eq:app:bg:solution}
\end{equation}
Positive definiteness also makes the quadratic term in
\Eqref{eq:app:bg:expanded-objective} strictly convex in $M$.
The stationary point is therefore the unique global minimizer.
Finally, substituting the eigendecomposition of $R$ into this solution
gives the gains in \Eqref{eq:bg:gains}.

\subsection{Proof of the pooling decomposition}
\label{app:pool:decomposition}

\begin{proof}[Proof of Proposition~\ref{prop:pool:decomposition}]
{Let $R_+$ and $R_-$ denote the pole covariances centered on their
respective means, $\delta=\mu_+-\mu_-$, and $p,q>0$ with $p+q=1$.
Write $R_{\mathrm{w}}=pR_++qR_-$,
$A=R_{\mathrm{w}}+\lambda I$, and
$\bar C=pC_\lambda(R_+)+qC_\lambda(R_-)$, using the same
$\lambda>0$ for all filters.} For finite fitting sets,
expectations are empirical averages normalized by $1/n_s$ within
each pole.

\paragraph{The pooled covariance.}
The pooled mean is $\mu=p\mu_++q\mu_-$, so
$\mu_+-\mu=q\delta$ and $\mu_--\mu=-p\delta$.
For the positive pole, write
$h-\mu=(h-\mu_+)+q\delta$.
Since $\mathbb{E}_+[h-\mu_+]=0$, expansion gives
\[
    \mathbb{E}_+[(h-\mu)(h-\mu)^\top]
    =R_++q^2\delta\delta^\top.
\]
The negative pole similarly gives $R_-+p^2\delta\delta^\top$.
Taking their weighted average yields
\begin{align}
    R_{\mathrm{pool}}
    &=pR_++qR_-+(pq^2+qp^2)\delta\delta^\top \nonumber\\
    &=R_{\mathrm{w}}+pq\delta\delta^\top.
    \label{eq:app:pool:covariance}
\end{align}

\paragraph{The mean update.}
The conceptor formula can be written as
$C_\lambda(R)=I-\lambda(R+\lambda I)^{-1}$.
Set $A=R_{\mathrm{w}}+\lambda I$ and $t=pq$.
Because $A\succ0$, the Sherman--Morrison identity gives
\begin{equation}
    (A+t\delta\delta^\top)^{-1}
    =A^{-1}
     -\frac{tA^{-1}\delta\delta^\top A^{-1}}
     {1+t\delta^\top A^{-1}\delta}.
    \label{eq:app:pool:inverse-update}
\end{equation}
Consequently,
\begin{align}
    C_\lambda(R_{\mathrm{pool}})-C_\lambda(R_{\mathrm{w}})
    &=\lambda\bigl[A^{-1}
      -(A+t\delta\delta^\top)^{-1}\bigr] \nonumber\\
    &=\frac{\lambda pq}
      {1+pq\delta^\top A^{-1}\delta}
      (A^{-1}\delta)(A^{-1}\delta)^\top
      =D_\mu.
    \label{eq:app:pool:mean-update}
\end{align}
The scalar multiplier is positive. Since $A$ is invertible,
$D_\mu$ is positive semidefinite, has rank one for $\delta\ne0$,
and is zero for $\delta=0$.
If $R_{\mathrm{w}}v_j=r_jv_j$, the $v_j$ coordinate of
$A^{-1}\delta$ is $(v_j^\top\delta)/(r_j+\lambda)$.
This gives the weighting described in the main text.

\paragraph{The covariance-shape update.}
Define
\[
    A_+=R_++\lambda I,\qquad
    A_-=R_-+\lambda I,\qquad
    H=R_+-R_-.
\]
Then $A=pA_++qA_-$, $A_+=A+qH$, and $A_-=A-pH$.
Using the inverse form of the conceptor gives
\begin{equation}
    J_R=C_\lambda(R_{\mathrm{w}})-\bar C
       =\lambda\bigl[pA_+^{-1}+qA_-^{-1}-A^{-1}\bigr].
    \label{eq:app:pool:shape-inverses}
\end{equation}
To obtain its sign and rank, we use the exact identity
\begin{equation}
    (A+E)^{-1}
    =A^{-1}-A^{-1}EA^{-1}
     +A^{-1}E(A+E)^{-1}EA^{-1}.
    \label{eq:app:pool:inverse-expansion}
\end{equation}
It follows by substituting
$(A+E)^{-1}=A^{-1}-(A+E)^{-1}EA^{-1}$
into
$(A+E)^{-1}=A^{-1}-A^{-1}E(A+E)^{-1}$.
Applying~\Eqref{eq:app:pool:inverse-expansion} to
$E=qH$ and $E=-pH$ gives
\begin{align*}
    A_+^{-1}
    &=A^{-1}-qA^{-1}HA^{-1}
      +q^2A^{-1}HA_+^{-1}HA^{-1},\\
    A_-^{-1}
    &=A^{-1}+pA^{-1}HA^{-1}
      +p^2A^{-1}HA_-^{-1}HA^{-1}.
\end{align*}
Weighting these equations by $p$ and $q$ cancels the terms that are
linear in $H$. Substitution into~\Eqref{eq:app:pool:shape-inverses} yields
\begin{equation}
    J_R=\lambda pq\,A^{-1}HBHA^{-1},
    \qquad B=qA_+^{-1}+pA_-^{-1}.
    \label{eq:app:pool:shape-factorization}
\end{equation}
Both $A$ and $B$ are positive definite. With
$K=B^{1/2}HA^{-1}$, the factorization becomes
$J_R=\lambda pq\,K^\top K\succeq0$.
The matrices on either side of $H$ in $K$ are invertible, giving
\[
    \operatorname{rank}(J_R)
    =\operatorname{rank}(K)
    =\operatorname{rank}(H).
\]
Thus $J_R=0$ exactly when $R_+=R_-$.
This argument allows arbitrary relative orientations of the two
covariances.

\paragraph{Combining the terms.}
The two updates satisfy
$C_\lambda(R_{\mathrm{pool}})=C_\lambda(R_{\mathrm{w}})+D_\mu$
and $C_\lambda(R_{\mathrm{w}})=\bar C+J_R$.
Substituting the second equality into the first proves~\Eqref{eq:pool:decomposition}.
\end{proof}

\subsection{Mean-only conceptors and the mean-difference axis}
\label{app:pool:rankone}

The standard conceptor spectral formula \citep{jaeger2014,jaeger2017}
has an exact mean-difference case.

\begin{proposition}[Mean-difference axis]
\label{prop:pool:rankone}
Let $\delta=\mu_+-\mu_-\ne0$ and $u=\delta/\|\delta\|_2$.
Fitting only the two equally weighted pole means about their midpoint
$c=(\mu_++\mu_-)/2$ gives
\begin{equation}
    C_{\mathrm{means}}
    =\frac{\|\delta\|_2^2}{\|\delta\|_2^2+4\lambda}\,uu^\top,
    \qquad \lambda>0.
    \label{eq:pool:meanonly}
\end{equation}
Its sole retained axis is the mean-difference line. When all examples
are fitted, the full conceptor remains rank one precisely when
within-pole variation is confined to this line.
\end{proposition}

The labels orient $u$ toward the positive mean; filtering scales each
input's coordinate along this axis by the regularized gain. Compressing
each pole to its mean gives one direction; variation outside that line
adds retained directions. 

We prove Proposition~\ref{prop:pool:rankone} using the spectral formula
in~\Eqref{eq:bg:gains}, following \citet{jaeger2014,jaeger2017}.
Throughout, $\delta=\mu_+-\mu_-\ne0$,
$u=\delta/\|\delta\|_2$, and $\lambda>0$.

\paragraph{Fitting the two means.}
Let the pole weights be $p,q>0$, with $p+q=1$.
Their pooled mean is $\mu=p\mu_++q\mu_-$, so centering gives
$x_+=\mu_+-\mu=q\delta$ and $x_-=\mu_--\mu=-p\delta$.
The resulting moment is
\begin{equation}
    R_{\mathrm{means}}
    =px_+x_+^\top+qx_-x_-^\top
    =pq\delta\delta^\top
    =pq\|\delta\|_2^2 uu^\top.
    \label{eq:app:pool:means-moment}
\end{equation}
It has one positive eigenvalue, $pq\|\delta\|_2^2$, along $u$.
The conceptor transforms this eigenvalue by $r\mapsto r/(r+\lambda)$,
giving
\begin{equation}
    C_\lambda(R_{\mathrm{means}})
    =\frac{pq\|\delta\|_2^2}{pq\|\delta\|_2^2+\lambda}\,uu^\top.
    \label{eq:app:pool:mean-only}
\end{equation}
Setting $p=q=1/2$ proves~\Eqref{eq:pool:meanonly}.
The same calculation applies to a presence--absence comparison after
replacing $\mu_-$ by $\mu_0$.

\paragraph{The positive-mean construction.}
For balanced weights, fit the single positive mean about the same
midpoint $c=(\mu_++\mu_-)/2$.
Its fitting vector is $\mu_+-c=\delta/2$, so its second moment is
$\delta\delta^\top/4$.
It therefore gives exactly the same conceptor as the balanced two-mean
fit at the same $\lambda$.
Both pole means are used to determine this reference.

\paragraph{Including within-pole variation.}
Let $R_{\mathrm{w}}=pR_++qR_-$ be the within-pole covariance in~\Eqref{eq:pool:covariance}, and write
$C_{\mathrm{pool}}=C_\lambda(R_{\mathrm{pool}})$.
The spectral transformation preserves zero and positive eigenvalues,
so $C_{\mathrm{pool}}$ and $R_{\mathrm{pool}}$ have the same range.
For any $z\in\mathbb{R}^d$,
\begin{equation}
    z^\top R_{\mathrm{pool}}z
    =z^\top R_{\mathrm{w}}z+pq(\delta^\top z)^2.
    \label{eq:app:pool:quadratic-form}
\end{equation}
Both terms are nonnegative. Their sum vanishes exactly when
$z\in\ker(R_{\mathrm{w}})\cap\delta^\perp$.
Taking orthogonal complements gives
\begin{equation}
    \operatorname{range}(C_{\mathrm{pool}})
    =\operatorname{range}(R_{\mathrm{w}})
     +\operatorname{span}(\delta).
    \label{eq:app:pool:range}
\end{equation}
Since $\delta\ne0$, this range is one-dimensional exactly when
$\operatorname{range}(R_{\mathrm{w}})\subseteq
\operatorname{span}(\delta)$.
This proves the full-filter rank-one condition stated in
Proposition~\ref{prop:pool:rankone}.

\subsection{Overlap and hard subspace removal}
\label{app:bg:overlap}

Let $U_A,U_B\in\mathbb{R}^{d\times k}$ have orthonormal columns spanning
two subspaces, and let $P_A=U_AU_A^\top$.
Define their overlap as
\begin{equation}
    \operatorname{Overlap}(A,B)
    =\frac{1}{k}\|U_A^\top U_B\|_F^2.
    \label{eq:app:bg:overlap}
\end{equation}
Removing subspace $A$ applies $I-P_A$.
For each unit column $u_j$ of $U_B$, orthogonality gives
$\|(I-P_A)u_j\|_2^2=1-\|U_A^\top u_j\|_2^2$.
Averaging over the $k$ columns yields
\begin{equation}
    \frac{1}{k}\|(I-P_A)U_B\|_F^2
    =1-\frac{1}{k}\|U_A^\top U_B\|_F^2
    =1-\operatorname{Overlap}(A,B).
    \label{eq:app:bg:removal}
\end{equation}
Overlap therefore measures the fraction of average squared length lost
from these axes under hard removal. The identity assumes orthogonal
projectors and equal weighting of the basis vectors. Soft conceptor
operations also depend on their gains. Effects on semantic content
require separate evaluation.

\section{Dataset details}
\label{app:data}

\subsection{Authored concept collections}
\label{app:data:authored}

\paragraph{BipolarConcepts.}
Each matched pair contains one sentence from each pole.
The inspected native inputs contain 100 pairs
each for Sentiment, Depression and Pol. Stance (600 sentence entries).
The depression counter-expressions include positive well-being.
The labels describe the content expressed in the sentences.
Table~\ref{tab:data:inventory} gives the inventory for both
authored concept collections.

\paragraph{ConceptVariants.}
Each topic is expressed in several formulations. For sentiment and
political leaning, descriptive-neutral text supplies a topic description;
attitudinal-neutral and undecided text supply alternative references.
The symptom comparisons distinguish ordinary functioning, explicit
symptom absence, flourishing, and recovery. All formulations of a topic
remain together when constructing evaluation folds. The static
comparator-reconstruction analysis excludes recovery because it
introduces a temporal change. There are 24 topics per concept, with one
sentence per formulation. 
\begin{table}[htbp]
\centering
\small
\setlength{\tabcolsep}{4pt}
\begin{tabularx}{\linewidth}{
    @{}ll>{\raggedright\arraybackslash}Xr@{}
}
\toprule
Collection & Concept & Formulations & Texts \\
\midrule
\textsc{BipolarConcepts}
 & Sentiment
 & Positive; negative
 & 200 \\
 & Pol. Stance
 & Left; right
 & 200 \\
 & Depression expression
 & Depression; counter-expressions
 & 200 \\
\midrule
\textsc{ConceptVariants}
 & Sentiment
 & Positive; negative; descriptive neutral; attitudinal neutral
 & 96 \\
 & Pol. Stance
 & Left; right; descriptive neutral; undecided
 & 96 \\
 & Depression expression
 & Depression; ordinary functioning; explicit absence; flourishing; recovery
 & 120 \\
\bottomrule
\end{tabularx}
\caption{Authored collections: formulations and sentence counts.}
\label{tab:data:inventory}
\end{table}

\subsection{ConceptGroups}
\label{app:data:conceptgroups}

The suite contains three collections with different grouping units.
Table~\ref{tab:data:lengths} gives their individual group sizes alongside
word-length statistics.

\paragraph{Sentiment.}
The texts come from UCI Sentiment Labelled Sentences
\citep{kotzias2015sentiment}. Amazon, IMDb, and Yelp define the three
groups. We retain the source positive/negative labels and use the review
domain to define evaluation folds.

\paragraph{Pol. Stance.}
The tweets come from SemEval/TweetEval
\citep{mohammad2016stance,barbieri2020tweeteval}. Preparation combines
the original training and test files and retains favor/against labels.
Removing four repeated texts from 998 packaged rows leaves 994 tweets
across abortion, atheism, climate change, feminism, and Hillary Clinton.
Evaluation uses custom held-topic folds. The labels express stance toward
a specified topic.

\paragraph{Depression.}
The authored first-person sentences span fourteen facets, including
sleep, energy, and interest. Each facet contains seven matched pairs
contrasting symptom expression with ordinary functioning or positive
well-being. Pair members remain together during sampling and splitting.
These sentences serve as linguistic stimuli.

\paragraph{Grouped evaluation.}
Every prepared domain, facet, or topic contains equal numbers of examples
from its two labels. Each collection is evaluated separately. One entire
group is excluded from fitting, and its labels are used to compute the
evaluation mean difference. Fitting weights source groups equally and
weights the two labels equally within each source group. Any selection of
a direction, gain, or layer uses source groups only. We repeat over the
held-out groups and average their scores equally.

\subsection{Public probing benchmarks}
\label{app:data:public}

Table~\ref{tab:data:public} records the targets and partitions used for
probing. The three sentiment benchmarks are compared with sentiment
spectra, Hyperpartisan News with political-leaning spectra, and the two
Reddit benchmarks with depression-expression spectra.

\begin{table}[htbp]
\centering
\small
\setlength{\tabcolsep}{4pt}
\begin{tabularx}{\linewidth}{@{}
    >{\raggedright\arraybackslash}p{0.38\linewidth}
    >{\raggedright\arraybackslash}X
    >{\raggedright\arraybackslash}p{0.24\linewidth}@{}
}
\toprule
Benchmark & Probe target & Training / evaluation \\
\midrule
SST-2 \citep{socher2013sst}
 & Positive / negative sentiment
 & Train / validation \\
Rotten Tomatoes \citep{pang2005rotten}
 & Positive / negative sentiment
 & Train / test \\
TweetEval sentiment \citep{barbieri2020tweeteval}
 & Positive / negative sentiment
 & Train / test \\
Hyperpartisan News \citep{kiesel2019hyperpartisan}
 & Hyperpartisan / mainstream
 & Train / validation \\
Reddit Depression \citep{pixel42depression}
 & Released binary proxy label
 & Train / test \\
Cross-cultural Reddit \citep{rai2025crosscultural}
 & Mental-health proxy label
 & Existing train / test \\
\bottomrule
\end{tabularx}
\caption{Public probing targets and recorded train/evaluation partitions.}
\label{tab:data:public}
\end{table}

TweetEval sentiment excludes neutral labels. Hyperpartisan News uses
its hyperpartisan/mainstream target, which differs from left/right
leaning. The Reddit labels are proxies and do not establish clinical
diagnoses. Cross-cultural Reddit uses pre-existing files described as a
user-level train/test split.

\subsection{Length statistics}
\label{app:data:lengths}

Word counts use whitespace-delimited units before tokenization.
Table~\ref{tab:data:lengths} reports mean, 95th-percentile, and maximum
word lengths over each complete prepared group or collection. \textsc{ConceptVariants}
additionally averages 63.0 Unicode characters per sentence.

\begin{table}[htbp]
\centering
\small
\setlength{\tabcolsep}{4pt}
\begin{tabularx}{\linewidth}{@{}lXrrrr@{}}
\toprule
Collection & Group & Texts & Mean words & 95th pct. & Max. \\
\midrule
\textsc{ConceptVariants}
 & All concepts & 312 & 10.0 & 14.0 & 16 \\
\midrule
\multicolumn{6}{@{}l}{\textsc{ConceptGroups}} \\
Sentiment
 & Amazon & 200 & 11.5 & 26.0 & 30 \\
 & IMDb & 600 & 14.3 & 33.0 & 61 \\
 & Yelp & 600 & 11.0 & 23.0 & 32 \\
Pol. Stance
 & Abortion & 240 & 16.4 & 24.0 & 30 \\
 & Atheism & 230 & 17.7 & 24.0 & 31 \\
 & Climate change & 48 & 16.0 & 22.3 & 24 \\
 & Feminism & 240 & 17.4 & 24.0 & 26 \\
 & Hillary Clinton & 236 & 14.9 & 23.0 & 25 \\
Depression
 & All 14 facets & 196 & 6.8 & 9.0 & 12 \\
\bottomrule
\end{tabularx}
\caption{Prepared sentence counts and word lengths; Depression aggregates fourteen equal-sized facets.}
\label{tab:data:lengths}
\end{table}

\begin{table}[htbp]
\centering
\small
\setlength{\tabcolsep}{5pt}
\begin{tabularx}{\linewidth}{@{}Xcc@{}}
\toprule
Collection & Gemma median / max. & Qwen median / max. \\
\midrule
\textsc{ConceptVariants}
 & 12 / 19 & 11 / 18 \\
\textsc{ConceptGroups}: Sentiment
 & 14 / 82 & 13 / 78 \\
\textsc{ConceptGroups}: Pol. Stance
 & 25 / 48 & 24 / 50 \\
\textsc{ConceptGroups}: Depression
 & 9 / 16 & 8 / 14 \\
\bottomrule
\end{tabularx}
\caption{Recorded token-count medians and maxima for the prepared geometry inputs.}
\label{tab:data:tokens}
\end{table}

Table~\ref{tab:data:tokens} reports token medians and maxima from the
Gemma~2 and Qwen~2.5 tokenizers. Inputs are plain text with special tokens,
without a chat template or truncation. 

\subsection{Sentence examples}
\label{app:data:examples}

Tables~\ref{tab:data:examples-bipolar}--\ref{tab:data:examples-stance}
illustrate every collection, comparator formulation, and evaluation group.
For Depression, the positive pole denotes symptom expression; authored
counterparts include ordinary functioning and positive well-being.
Recovery appears in the inventory but is excluded from the static
comparator and amplification analyses. Pol. Stance denotes left/right
policy contrasts in the authored collections and favor/against positions
on specified topics in \textsc{ConceptGroups}.
Review and tweet examples placed in the same row are independent texts,
not matched sentence pairs.

% Real text examples; provenance is in revision_notes/EXAMPLES_PROVENANCE.md.
% Generated from real source rows by revision_notes/build_dataset_audit.py.
\begin{table}[t]
\centering
\small
\setlength{\tabcolsep}{4pt}
\renewcommand{\arraystretch}{1.08}
\begin{tabularx}{\linewidth}{@{}>{\raggedright\arraybackslash}p{0.15\linewidth}>{\raggedright\arraybackslash}X>{\raggedright\arraybackslash}X@{}}
\toprule
Concept & Recorded positive pole & Recorded negative pole \\
\midrule
Sentiment & I feel great today.\; & I feel bad today.\; \\
Pol. Stance & I support student debt relief.\; & I oppose student debt cancellation.\;\\
Depression & Nothing interests me.\; & Everything interests me.\;\\
\bottomrule
\end{tabularx}
\caption{Matched \textsc{BipolarConcepts} examples; positive denotes sentiment, left-leaning policy, or depression expression.}
\label{tab:data:examples-bipolar}
\end{table}

% Generated from real source rows by revision_notes/build_dataset_audit.py.
\begin{table}[t]
\centering
\small
\setlength{\tabcolsep}{4pt}
\renewcommand{\arraystretch}{1.08}
\begin{tabularx}{\linewidth}{@{}>{\raggedright\arraybackslash}p{0.15\linewidth}>{\raggedright\arraybackslash}p{0.24\linewidth}>{\raggedright\arraybackslash}X@{}}
\toprule
Concept & Formulation & Matched-topic example \\
\midrule
Sentiment & Positive & I found the film wonderfully engaging.\; \\
 & Negative & I found the film painfully dull.\; \\
 & Descriptive neutral & I watched the film in the evening.\; \\
 & Attitudinal neutral & I had no strong feelings about the film.\; \\
Pol. Stance & Left & Regarding health insurance policy, I favor expanding publicly funded health coverage.\; \\
 & Right & Regarding health insurance policy, I favor relying more on privately purchased health coverage.\; \\
 & Descriptive neutral & Regarding health insurance policy, different proposals are being discussed by the public.\; \\
 & Undecided & Regarding health insurance policy, I have not settled on a policy preference.\; \\
Depression & Symptom expression & My usual hobbies have lost all their appeal.\; \\
 & Ordinary functioning & My usual hobbies interest me about as much as before.\; \\
 & Explicit absence & My usual hobbies have not lost their ordinary appeal.\; \\
 & Flourishing & My usual hobbies fill me with excitement and delight.\; \\
 & Recovery & My usual hobbies interest me again after a period of losing their appeal.\; \\
\bottomrule
\end{tabularx}
\caption{Every \textsc{ConceptVariants} formulation, illustrated by one topic per concept; recovery is excluded from static analyses.}
\label{tab:data:examples-variants}
\end{table}

% Generated from real source rows by revision_notes/build_dataset_audit.py.
\begin{table}[t]
\centering
\small
\setlength{\tabcolsep}{4pt}
\renewcommand{\arraystretch}{1.08}
\begin{tabularx}{\linewidth}{@{}>{\raggedright\arraybackslash}p{0.16\linewidth}>{\raggedright\arraybackslash}X>{\raggedright\arraybackslash}X@{}}
\toprule
Review domain & Positive & Negative \\
\midrule
Amazon & Battery life is also great!\; & Awkward to use and unreliable.\; \\
IMDb & This is such a fun and funny movie.\; & The problem was the script.\; \\
Yelp & The atmosphere here is fun.\; & The manager was the worst.\; \\
\bottomrule
\end{tabularx}
\caption{\textsc{ConceptGroups}: positive and negative Sentiment examples from each review domain.}
\label{tab:data:examples-reviews}
\end{table}

% Generated from real source rows by revision_notes/build_dataset_audit.py.
\begin{table}[t]
\centering
\small
\setlength{\tabcolsep}{4pt}
\renewcommand{\arraystretch}{1.08}
\begin{tabularx}{\linewidth}{@{}>{\raggedright\arraybackslash}p{0.16\linewidth}>{\raggedright\arraybackslash}X>{\raggedright\arraybackslash}X@{}}
\toprule
Facet & Symptom expression & Ordinary / positive counterpart \\
\midrule
Appetite & I have no appetite and food feels like a chore.\; & I enjoy my meals and eat well.\; \\
Burden & I feel like a burden to everyone around me.\; & I feel valued by the people around me.\; \\
Concentration & I can't concentrate on anything anymore.\; & I concentrate easily on what I'm doing.\; \\
Energy & I feel drained and tired all the time.\; & I feel energetic and lively all the time.\; \\
Hopelessness & The future looks completely bleak to me.\; & I look forward to the future with hope.\;\\
Interest & Nothing interests me anymore.\; & I'm genuinely interested in lots of things.\; \\
Irritability & Every little thing makes me snap.\;& Little things don't get to me.\; \\
Mood & I feel sad and down most of the time.\; & I feel content most of the time.\; \\
Psychomotor & I move and speak as if in slow motion.\; & I move and speak at my normal pace.\; \\
Sleep & I lie awake for hours every night.\; & I fall asleep easily every night.\; \\
Suicidality & Sometimes I wish I just wouldn't wake up.\;& I'm glad to wake up to each day.\;\\
Tearfulness & I cry all the time, often for no reason.\; & I rarely feel the urge to cry.\; \\
Withdrawal & I've cut myself off from everyone I know.\; & I stay close to the people I know.\; \\
Worthlessness & I feel like a worthless failure.\; & I feel capable and worthwhile.\; \\
\bottomrule
\end{tabularx}
\caption{\textsc{ConceptGroups}: matched Depression expressions from all fourteen facets.}
\label{tab:data:examples-facets}
\end{table}

% Generated from real source rows by revision_notes/build_dataset_audit.py.
\begin{table}[t]
\centering
\small
\setlength{\tabcolsep}{4pt}
\renewcommand{\arraystretch}{1.08}
\begin{tabularx}{\linewidth}{@{}>{\raggedright\arraybackslash}p{0.16\linewidth}>{\raggedright\arraybackslash}X>{\raggedright\arraybackslash}X@{}}
\toprule
Topic & Favor & Against \\
\midrule
Abortion & So prolifers are against health rights?\; & \#ProLifeYouth know that life begins at conception.\; \\
Atheism & god of the gaps is not evidence \#next\; & God is so good! All the praises go up to Him!\; \\
Climate change & RT @user 700 people dead in Pakistan due to heatwave.\; & @user ice cracking in the summer?? SO ALARMING. \#climatechangehoax\; \\
Feminism & FINALLY A WOMEN RUNNING FOR PRESIDENT\; & Feminism is nothing more than a cult that closes ranks whenever someone dares question them.\; \\
Hillary Clinton & RT @user Hillary for the win\; & Hillary Clinton for Ambassador to ISIS.\; \\
\bottomrule
\end{tabularx}
\caption{\textsc{ConceptGroups}: favor/against examples for all five Pol. Stance topics.}
\label{tab:data:examples-stance}
\end{table}

\clearpage

% Append after the existing mathematical and data appendices.
% The parent document supplies \appendix once.
\section{Additional experimental details and results}
\label{app:c1}

\subsection{Feature sites, fitting, and uncertainty}
\label{app:c1:protocol}

Table~\ref{tab:extension:models} lists the five models, hidden dimensions,
and zero-based midpoint/final geometry sites. Mean pooling uses
nonpadding tokens; last-token controls use the last nonpadding token.
Layer plots display one-based decoder-layer numbers. 

\begin{table}[htbp]
\centering\small
\setlength{\tabcolsep}{4pt}
\begin{tabular}{@{}lrrrr@{}}
\toprule
Model & Blocks & $d$ & Midpoint & Final \\
\midrule
Gemma-2-2B-IT & 26 & 2,304 & 13 & 25 \\
Gemma-2-9B-IT & 42 & 3,584 & 21 & 41 \\
Qwen-2.5-3B-Instruct & 36 & 2,048 & 18 & 35 \\
Llama-3.1-8B-Instruct & 32 & 4,096 & 16 & 31 \\
OLMo-3-7B-Instruct & 32 & 4,096 & 16 & 31 \\
\bottomrule
\end{tabular}
\caption{Five-model geometry sites (zero-based raw blocks). }
\label{tab:extension:models}
\end{table}

For \textsc{ConceptGroups}, one complete group is held out in each fit.
The three collections supply three review domains, fourteen symptom
facets, and five stance topics, yielding 110 held-model/group fits per
feature site. Source groups receive equal weight, with equal pole
weights within each group. Empirical covariances use population
normalization. Held groups contribute only to evaluation.
Any selection of a direction, gain, or readout penalty uses source groups.
The source weights, evaluated rows, and numerical settings are fixed
within each comparison.

The reference experiment uses 2,000 complete-pipeline bootstrap refits,
as detailed in Appendix~\ref{app:c1:reference}. Archived coverage
intervals instead resample held-group scores 10,000 times while keeping
fits fixed; component intervals use 2,000 such draws. These intervals
condition on the fitted folds. There are only three review-domain
resampling units.

\subsection{Pooling controls and the original fitting collections}
\label{app:c1:pooling}

\paragraph{Matched counts.}
Within each source group, a deterministic hash ordering selects a bank
of $B$ rows per pole, with $B$ the largest eligible even count up to 32.
Each fit receives $B$ rows per group: half from each pole for pooling,
or all from one pole otherwise. Total fitting budgets are 64 rows for
Sentiment, 78 for Depression, and 96 or 128 for Pol. Stance.
The shared centroid/calibration bank has twice the fitting count.
Authored pair units remain intact.

Own-centered fits subtract their respective fitting means. Shared-center
fits subtract the same balanced source-bank center. Ranks 1, 10, and 20
are evaluated at the midpoint/mean-token sites. Table~\ref{tab:c1:centering}
reports rank twenty. Pooled rank-one support exceeds the count-matched
mean line in nine of fifteen conditions: all five Pol. Stance conditions
and four of five Sentiment conditions. It falls below the mean line for
OLMo Sentiment and all five Depression conditions.

\begin{figure}[!htbp]
\centering
\includegraphics[width=0.95\linewidth]{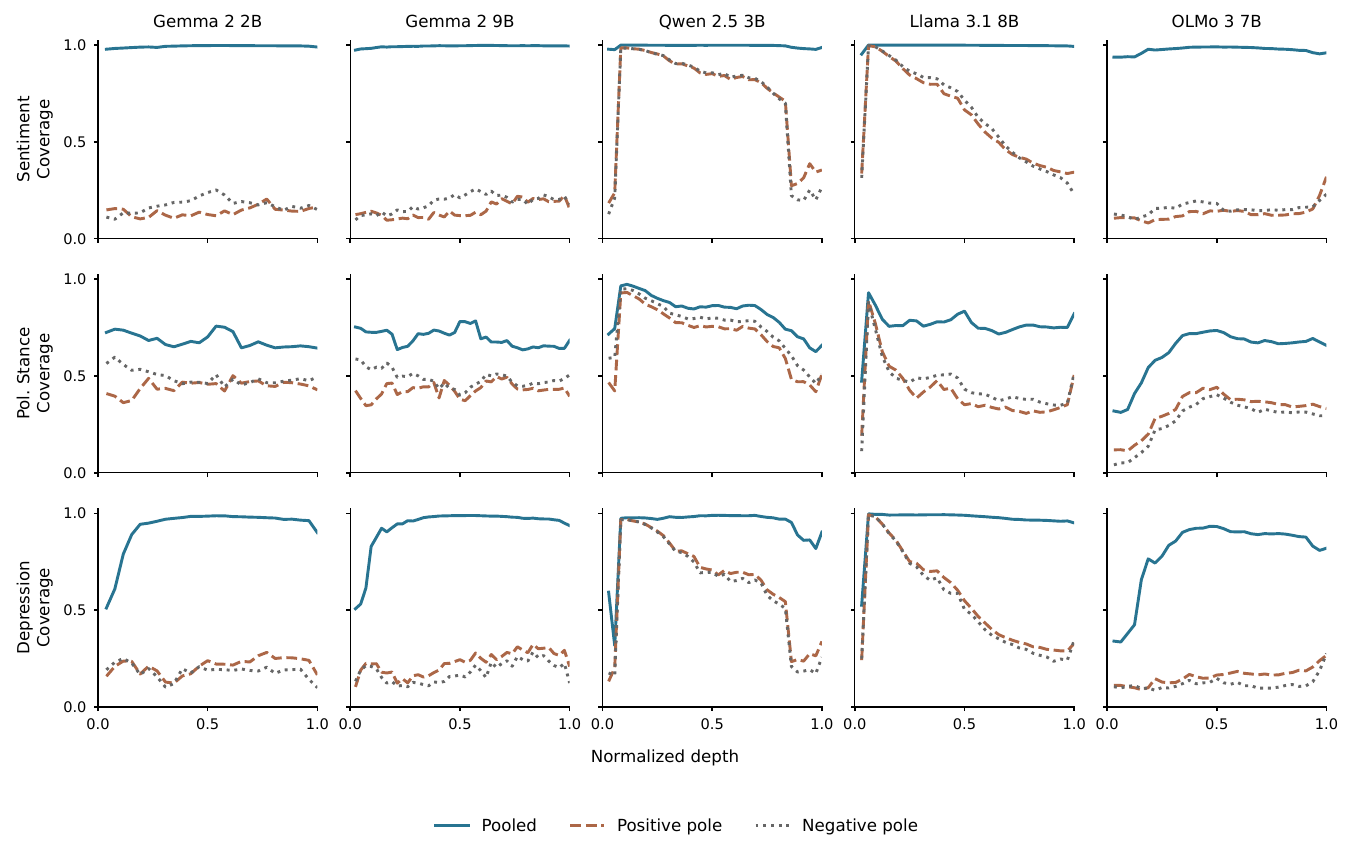}
\caption{\textbf{Original bipolar capture.} Leading-twenty coverage of each fitting collection's own mean difference across layers; these are training contrasts.}
\label{fig:c1:original}
\end{figure}

\paragraph{\textsc{BipolarConcepts}}
  Figure~\ref{fig:c1:original} reports leading-twenty capture of each fitting
  collection's own mean difference. Pooled capture has the highest layer
  average in all fifteen model--concept pairs; Sentiment averages range
  from 97.4\% to 99.7\%. These training contrasts describe fitted geometry
  and are distinct from the held-group evaluations.

\subsection{Complete decomposition ablations and gain controls}
\label{app:c1:ablation}

\paragraph{Shared regularization.}
All four operators use full nonzero fitted supports and share the
source fold's regularization:
\[
    \lambda=\tau\,\operatorname{tr}(R_{\mathrm{pool}})/d,
    \qquad \tau\in\{0.1,1,10\},
\]
with $\tau=1$ as the primary scale.
Tables~\ref{tab:c1:ablation:gemma2_2b_it}--\ref{tab:c1:ablation:olmo3_7b_it}
report both mean-token sites and all scales. The full pooled filter has
lower group-averaged reconstruction error than each ablation in all 90
model/data/site/scale settings: 270 paired comparisons, with full-minus-
ablated errors from $-0.224842$ to $-0.000283$.

Every operator receives the held contrast, which identity reconstructs
with zero error. These comparisons concern the fitted filter family;
positive semidefiniteness alone does not guarantee the observed ordering
for every vector.

\begin{table}[!htbp]
\centering
\small
\setlength{\tabcolsep}{4pt}
\begin{tabular}{@{}llrrrrr@{}}
\toprule
Data & Site & $\tau$ & $\bar C$ & $+D_\mu$ & $+J_R$ & Full \\
\midrule
Sentiment & Mid & 0.1 & 0.1344 & 0.1052 & 0.0747 & \textbf{0.0536} \\
 & Mid & 1 & 0.2489 & 0.1466 & 0.1679 & \textbf{0.0984} \\
 & Mid & 10 & 0.4637 & 0.2689 & 0.4284 & \textbf{0.2460} \\
 & Final & 0.1 & 0.1964 & 0.1797 & 0.1262 & \textbf{0.1139} \\
 & Final & 1 & 0.2792 & 0.2347 & 0.2166 & \textbf{0.1855} \\
 & Final & 10 & 0.4805 & 0.3584 & 0.4376 & \textbf{0.3294} \\
\midrule
Pol. Stance & Mid & 0.1 & 0.1614 & 0.1605 & 0.1147 & \textbf{0.1139} \\
 & Mid & 1 & 0.2088 & 0.2074 & 0.1786 & \textbf{0.1774} \\
 & Mid & 10 & 0.3539 & 0.3504 & 0.3346 & \textbf{0.3314} \\
 & Final & 0.1 & 0.1626 & 0.1619 & 0.1140 & \textbf{0.1135} \\
 & Final & 1 & 0.2087 & 0.2078 & 0.1765 & \textbf{0.1758} \\
 & Final & 10 & 0.3532 & 0.3510 & 0.3307 & \textbf{0.3287} \\
\midrule
Depression & Mid & 0.1 & 0.3111 & 0.2577 & 0.2114 & \textbf{0.1619} \\
 & Mid & 1 & 0.3265 & 0.2529 & 0.2300 & \textbf{0.1732} \\
 & Mid & 10 & 0.4068 & 0.2801 & 0.3461 & \textbf{0.2396} \\
 & Final & 0.1 & 0.4228 & 0.3748 & 0.3255 & \textbf{0.2792} \\
 & Final & 1 & 0.4321 & 0.3745 & 0.3365 & \textbf{0.2876} \\
 & Final & 10 & 0.4967 & 0.4102 & 0.4315 & \textbf{0.3585} \\
\bottomrule
\end{tabular}
\caption{\textbf{Gemma-2B reconstruction ablations.} Lower normalized error is better. Updates augment $\bar C$ at shared $\lambda=\tau\,\operatorname{tr}(R_{\mathrm{pool}})/d$; bold marks the row minimum.}
\label{tab:c1:ablation:gemma2_2b_it}
\end{table}

\begin{table}[!htbp]
\centering
\small
\setlength{\tabcolsep}{4pt}
\begin{tabular}{@{}llrrrrr@{}}
\toprule
Data & Site & $\tau$ & $\bar C$ & $+D_\mu$ & $+J_R$ & Full \\
\midrule
Sentiment & Mid & 0.1 & 0.1573 & 0.1262 & 0.0987 & \textbf{0.0756} \\
 & Mid & 1 & 0.2573 & 0.1626 & 0.1878 & \textbf{0.1214} \\
 & Mid & 10 & 0.4398 & 0.2722 & 0.4127 & \textbf{0.2545} \\
 & Final & 0.1 & 0.2506 & 0.2315 & 0.1750 & \textbf{0.1597} \\
 & Final & 1 & 0.3274 & 0.2820 & 0.2620 & \textbf{0.2298} \\
 & Final & 10 & 0.5093 & 0.3922 & 0.4666 & \textbf{0.3640} \\
\midrule
Pol. Stance & Mid & 0.1 & 0.2117 & 0.2101 & 0.1634 & \textbf{0.1620} \\
 & Mid & 1 & 0.2491 & 0.2460 & 0.2140 & \textbf{0.2116} \\
 & Mid & 10 & 0.3791 & 0.3718 & 0.3592 & \textbf{0.3526} \\
 & Final & 0.1 & 0.2450 & 0.2440 & 0.1883 & \textbf{0.1874} \\
 & Final & 1 & 0.2746 & 0.2732 & 0.2301 & \textbf{0.2290} \\
 & Final & 10 & 0.3952 & 0.3921 & 0.3697 & \textbf{0.3671} \\
\midrule
Depression & Mid & 0.1 & 0.2840 & 0.2331 & 0.1985 & \textbf{0.1506} \\
 & Mid & 1 & 0.2948 & 0.2281 & 0.2119 & \textbf{0.1587} \\
 & Mid & 10 & 0.3554 & 0.2458 & 0.3018 & \textbf{0.2099} \\
 & Final & 0.1 & 0.3604 & 0.3124 & 0.2858 & \textbf{0.2389} \\
 & Final & 1 & 0.3660 & 0.3110 & 0.2915 & \textbf{0.2431} \\
 & Final & 10 & 0.4069 & 0.3295 & 0.3529 & \textbf{0.2874} \\
\bottomrule
\end{tabular}
\caption{\textbf{Gemma-9B reconstruction ablations.} Lower normalized error is better. Updates augment $\bar C$ at shared $\lambda=\tau\,\operatorname{tr}(R_{\mathrm{pool}})/d$; bold marks the row minimum.}
\label{tab:c1:ablation:gemma2_9b_it}
\end{table}

\begin{table}[!htbp]
\centering
\small
\setlength{\tabcolsep}{4pt}
\begin{tabular}{@{}llrrrrr@{}}
\toprule
Data & Site & $\tau$ & $\bar C$ & $+D_\mu$ & $+J_R$ & Full \\
\midrule
Sentiment & Mid & 0.1 & 0.0291 & 0.0182 & 0.0218 & \textbf{0.0137} \\
 & Mid & 1 & 0.0584 & 0.0411 & 0.0552 & \textbf{0.0388} \\
 & Mid & 10 & 0.0783 & 0.0733 & 0.0780 & \textbf{0.0730} \\
 & Final & 0.1 & 0.1675 & 0.1545 & 0.1036 & \textbf{0.0937} \\
 & Final & 1 & 0.2213 & 0.1918 & 0.1679 & \textbf{0.1461} \\
 & Final & 10 & 0.3924 & 0.3040 & 0.3422 & \textbf{0.2678} \\
\midrule
Pol. Stance & Mid & 0.1 & 0.0920 & 0.0914 & 0.0748 & \textbf{0.0744} \\
 & Mid & 1 & 0.1650 & 0.1639 & 0.1550 & \textbf{0.1540} \\
 & Mid & 10 & 0.2882 & 0.2867 & 0.2837 & \textbf{0.2822} \\
 & Final & 0.1 & 0.1702 & 0.1698 & 0.1214 & \textbf{0.1211} \\
 & Final & 1 & 0.2142 & 0.2133 & 0.1831 & \textbf{0.1824} \\
 & Final & 10 & 0.3541 & 0.3517 & 0.3358 & \textbf{0.3336} \\
\midrule
Depression & Mid & 0.1 & 0.0717 & 0.0591 & 0.0502 & \textbf{0.0405} \\
 & Mid & 1 & 0.0870 & 0.0649 & 0.0733 & \textbf{0.0551} \\
 & Mid & 10 & 0.1158 & 0.0952 & 0.1138 & \textbf{0.0936} \\
 & Final & 0.1 & 0.4732 & 0.4164 & 0.3633 & \textbf{0.3096} \\
 & Final & 1 & 0.4851 & 0.4140 & 0.3787 & \textbf{0.3202} \\
 & Final & 10 & 0.5636 & 0.4577 & 0.4936 & \textbf{0.4034} \\
\bottomrule
\end{tabular}
\caption{\textbf{Qwen-3B reconstruction ablations.} Lower normalized error is better. Updates augment $\bar C$ at shared $\lambda=\tau\,\operatorname{tr}(R_{\mathrm{pool}})/d$; bold marks the row minimum.}
\label{tab:c1:ablation:qwen25_3b_it}
\end{table}

\begin{table}[!htbp]
\centering
\footnotesize
\setlength{\tabcolsep}{3pt}
\begin{tabular}{@{}llrrrrr@{}}
\toprule
Data & Site & $\tau$ & $\bar C$ & $+D_\mu$ & $+J_R$ & Pooled \\
\midrule
Sentiment & Mid & 0.1 & 0.1011 & 0.0786 & 0.0695 & \textbf{0.0540} \\
Sentiment & Mid & 1 & 0.1687 & 0.1080 & 0.1437 & \textbf{0.0931} \\
Sentiment & Mid & 10 & 0.2460 & 0.1980 & 0.2420 & \textbf{0.1947} \\
Sentiment & Final & 0.1 & 0.1698 & 0.1567 & 0.1096 & \textbf{0.0982} \\
Sentiment & Final & 1 & 0.1985 & 0.1765 & 0.1436 & \textbf{0.1274} \\
Sentiment & Final & 10 & 0.3237 & 0.2511 & 0.2677 & \textbf{0.2126} \\
Pol. Stance & Mid & 0.1 & 0.1998 & 0.1984 & 0.1595 & \textbf{0.1583} \\
Pol. Stance & Mid & 1 & 0.2446 & 0.2416 & 0.2192 & \textbf{0.2169} \\
Pol. Stance & Mid & 10 & 0.3647 & 0.3596 & 0.3514 & \textbf{0.3466} \\
Pol. Stance & Final & 0.1 & 0.1597 & 0.1589 & 0.1206 & \textbf{0.1199} \\
Pol. Stance & Final & 1 & 0.1820 & 0.1809 & 0.1520 & \textbf{0.1511} \\
Pol. Stance & Final & 10 & 0.2742 & 0.2721 & 0.2565 & \textbf{0.2545} \\
Depression & Mid & 0.1 & 0.1810 & 0.1507 & 0.1340 & \textbf{0.1063} \\
Depression & Mid & 1 & 0.1905 & 0.1484 & 0.1478 & \textbf{0.1144} \\
Depression & Mid & 10 & 0.2319 & 0.1712 & 0.2139 & \textbf{0.1580} \\
Depression & Final & 0.1 & 0.4407 & 0.3810 & 0.3306 & \textbf{0.2722} \\
Depression & Final & 1 & 0.4462 & 0.3785 & 0.3351 & \textbf{0.2752} \\
Depression & Final & 10 & 0.4936 & 0.3938 & 0.3991 & \textbf{0.3188} \\
\bottomrule
\end{tabular}
\caption{Llama-3.1-8B: held-contrast reconstruction error, averaged equally over held groups; lower is better.}
\label{tab:c1:ablation:llama31_8b_it}
\end{table}

\begin{table}[!htbp]
\centering
\footnotesize
\setlength{\tabcolsep}{3pt}
\begin{tabular}{@{}llrrrrr@{}}
\toprule
Data & Site & $\tau$ & $\bar C$ & $+D_\mu$ & $+J_R$ & Pooled \\
\midrule
Sentiment & Mid & 0.1 & 0.3104 & 0.2661 & 0.1887 & \textbf{0.1477} \\
Sentiment & Mid & 1 & 0.3352 & 0.2714 & 0.2136 & \textbf{0.1654} \\
Sentiment & Mid & 10 & 0.4719 & 0.3207 & 0.3575 & \textbf{0.2470} \\
Sentiment & Final & 0.1 & 0.2560 & 0.2346 & 0.1773 & \textbf{0.1581} \\
Sentiment & Final & 1 & 0.2799 & 0.2479 & 0.2078 & \textbf{0.1830} \\
Sentiment & Final & 10 & 0.4007 & 0.3250 & 0.3413 & \textbf{0.2814} \\
Pol. Stance & Mid & 0.1 & 0.3100 & 0.3090 & 0.2508 & \textbf{0.2498} \\
Pol. Stance & Mid & 1 & 0.3267 & 0.3256 & 0.2717 & \textbf{0.2707} \\
Pol. Stance & Mid & 10 & 0.4228 & 0.4211 & 0.3905 & \textbf{0.3890} \\
Pol. Stance & Final & 0.1 & 0.2571 & 0.2558 & 0.2044 & \textbf{0.2031} \\
Pol. Stance & Final & 1 & 0.2757 & 0.2742 & 0.2289 & \textbf{0.2275} \\
Pol. Stance & Final & 10 & 0.3701 & 0.3680 & 0.3426 & \textbf{0.3408} \\
Depression & Mid & 0.1 & 0.6674 & 0.5932 & 0.5519 & \textbf{0.4786} \\
Depression & Mid & 1 & 0.6709 & 0.5897 & 0.5539 & \textbf{0.4799} \\
Depression & Mid & 10 & 0.7018 & 0.5928 & 0.5988 & \textbf{0.5101} \\
Depression & Final & 0.1 & 0.5666 & 0.5165 & 0.4729 & \textbf{0.4237} \\
Depression & Final & 1 & 0.5703 & 0.5152 & 0.4753 & \textbf{0.4253} \\
Depression & Final & 10 & 0.6040 & 0.5312 & 0.5202 & \textbf{0.4597} \\
\bottomrule
\end{tabular}
\caption{OLMo-3-7B: held-contrast reconstruction error, averaged equally over held groups; lower is better.}
\label{tab:c1:ablation:olmo3_7b_it}
\end{table}

% Add within Appendix C.3, after the five original ablation tables and
% before \paragraph{Predicting sentence labels.}
\clearpage
\paragraph{Comparing directions at equal rank and trace.}
In a follow-up control, we replace each source-fitted operator
$M\in\mathcal M=\{\bar C,\bar C+D_\mu,\bar C+J_R,C_{\mathrm{pool}}\}$
by the orthogonal projector $P_{20}(M)=U_{20}(M)U_{20}(M)^\top$
onto its leading twenty eigenvectors. All fits have at least twenty
nonzero modes. Every projector therefore has rank twenty, trace twenty,
and twenty unit gains, with directions selected from source data alone.
These are hard projectors, rather than the soft filters $C_{20}$.
We retain the same held-group folds, sites, and shrinkage scales, and
average held groups equally. Coverage is measured by
\eqref{eq:data:error}; projector reconstruction error equals one minus
coverage.

The pooled projector has higher coverage in 251 of 270 comparisons
(Table~\ref{tab:c1:ablation:rank20}). At the primary midpoint setting
with $\tau=1$, it wins 41 of 45 comparisons and has the highest coverage
in 12 of 15 model/data settings. These counts describe the evaluated
grid, not independent replications; the pooled and within-covariance
projectors are unchanged across $\tau$.

\begin{table}[!htbp]
\centering
\small
\setlength{\tabcolsep}{6pt}
\begin{tabular}{@{}lrrr@{}}
\toprule
Ablated projector & Pooled wins & Pooled losses & Mean gain (pp) \\
\midrule
$P_{20}(\bar C)$ & 84 & 6 & $+14.56$ \\
$P_{20}(\bar C+D_\mu)$ & 77 & 13 & $+7.26$ \\
$P_{20}(\bar C+J_R)$ & 90 & 0 & $+10.10$ \\
\midrule
All comparisons & 251 & 19 & $+10.64$ \\
\bottomrule
\end{tabular}
\caption{Pooled versus ablated rank-20 projector coverage across five
models, three datasets, two sites, and three scales. Gains are
pooled-minus-ablation coverage in percentage points. Each ablation has
90 comparisons; no ties occur within $10^{-10}$.}
\label{tab:c1:ablation:rank20}
\end{table}

The advantage is not guaranteed. Pooling prioritizes total variance,
so within-pole variation can enter the leading twenty directions and
displace a direction aligned with the held contrast, even when the
source and held mean contrasts agree.

\paragraph{Equal trace for the full soft filters.}
As a separate control, we uniformly rescale the full operators using
only their source-fitted traces:
\[
    t_* = \min_{M\in\mathcal M}\operatorname{tr}(M),
    \qquad \widetilde M = \frac{t_*}{\operatorname{tr}(M)}M.
\]
This feasible budget only attenuates gains. After matching within each
fold and averaging held groups equally, the pooled filter has lower
reconstruction error in 137 comparisons and higher error in 133.
The projector result therefore supports an advantage in leading-subspace
coverage, without implying a universal reconstruction advantage for the
full soft filters at equal trace.

\paragraph{Predicting sentence labels.}
We test whether better contrast reconstruction also improves pole
prediction. A regularized linear classifier uses either the rank-one
mean-separation component $D_\mu$ or the pooled conceptor's leading
twenty modes. For each held-out group, the remaining groups supply
training, and source-group validation selects regularization.
Balanced accuracy measures the
average accuracy across the two poles on individual held-out sentences.

Mean-component features outperform pooled features in 25 of 30
model/data/site conditions using mean-token activations, with two ties
and three losses (Table~\ref{tab:c1:controls}). The full filter's
reconstruction advantage therefore does not imply better pole prediction. These results motivate different criteria for probing and
concept-preserving steering: a probe needs to distinguish the poles,
while an intervention may also need to preserve variation within
each pole.

\paragraph{Stabilizing mean differences estimated from few sentences.}
We test whether filtering stabilizes mean differences estimated from
few sentences. Within each held-out group, two disjoint samples,
each containing both poles, give $\widehat\delta_a$ and
$\widehat\delta_b$. Matched pairs stay in the same sample. We measure
agreement by filtering each estimate and comparing it with the other:
\[
    L(M)=\tfrac12\bigl(
    \|M\widehat\delta_a-\widehat\delta_b\|_2^2+
    \|M\widehat\delta_b-\widehat\delta_a\|_2^2\bigr).
\]
Both estimates are noisy. We average thirty splits per group, normalize
by that group's average loss with $M=I$, and average groups equally.
Loss below one improves on leaving the estimates unchanged.

Hard projection $P_{20}$ and the conceptor $C_{20}$ share twenty
source-fitted axes; only the conceptor shrinks retained coordinates.
Source-group validation selects its shrinkage scale from
$\{10^{-5},10^{-4},10^{-3},10^{-2},.1,1\}$.

With one sentence per pole in each sample, $C_{20}$ has lower loss than
$P_{20}$ in all 60 model/data/site/pooling conditions: for Gemma-2B
Sentiment, normalized loss decreases from
0.950 to 0.600 (Table~\ref{tab:c1:controls}).
This shrinkage benefit depends on sample size. With the largest
available Sentiment samples, identity has lower loss than conceptor
filtering in all twenty conditions. Shrinkage can suppress sampling
noise but also attenuate useful components of a well-estimated contrast.

\begin{table}[!htbp]
\centering
\footnotesize
\setlength{\tabcolsep}{3pt}
\begin{tabular}{@{}llrrrr@{}}
\toprule
& & \multicolumn{2}{c}{Denoising loss $\downarrow$} & \multicolumn{2}{c}{Readout BA (\%) $\uparrow$} \\
\cmidrule(lr){3-4}\cmidrule(l){5-6}
Model & Data & Hard $P_{20}$ & Conceptor $C_{20}$ & Mean update & Pooled \\
\midrule
Gemma-2B & Sentiment & 0.950 & \textbf{0.600} & 94.6 & \textbf{95.4} \\
 & Pol. Stance & 0.811 & \textbf{0.553} & \textbf{58.2} & 57.3 \\
 & Depression & 0.845 & \textbf{0.638} & \textbf{98.5} & 96.9 \\
\midrule
Gemma-9B & Sentiment & 0.964 & \textbf{0.608} & \textbf{95.0} & 94.7 \\
 & Pol. Stance & 0.833 & \textbf{0.567} & \textbf{64.3} & 59.5 \\
 & Depression & 0.878 & \textbf{0.646} & \textbf{99.0} & 98.0 \\
\midrule
Qwen-3B & Sentiment & 0.996 & \textbf{0.615} & 87.1 & \textbf{87.5} \\
 & Pol. Stance & 0.955 & \textbf{0.619} & \textbf{57.5} & 54.1 \\
 & Depression & 0.971 & \textbf{0.630} & \textbf{94.9} & \textbf{94.9} \\
\midrule
Llama-3.1-8B & Sentiment & 0.988 & \textbf{0.610} & 92.7 & \textbf{93.4} \\
 & Pol. Stance & 0.907 & \textbf{0.599} & \textbf{64.8} & 59.4 \\
 & Depression & 0.931 & \textbf{0.626} & \textbf{97.4} & \textbf{97.4} \\
\midrule
OLMo-3-7B & Sentiment & 0.604 & \textbf{0.514} & \textbf{93.2} & 91.7 \\
 & Pol. Stance & 0.615 & \textbf{0.520} & \textbf{61.1} & 58.4 \\
 & Depression & 0.599 & \textbf{0.559} & \textbf{97.4} & 95.4 \\
\bottomrule
\end{tabular}
\caption{Midpoint controls. One-sentence-per-pole loss is normalized to identity ($1$); hard and conceptor filters share twenty axes. Pole-prediction balanced accuracy (BA) uses source-tuned readouts.}
\label{tab:c1:controls}
\end{table}

\subsection{Layer coverage}
\label{app:c1:layers}

\begin{figure}[!htbp]
      \centering
      \includegraphics[width=\linewidth]{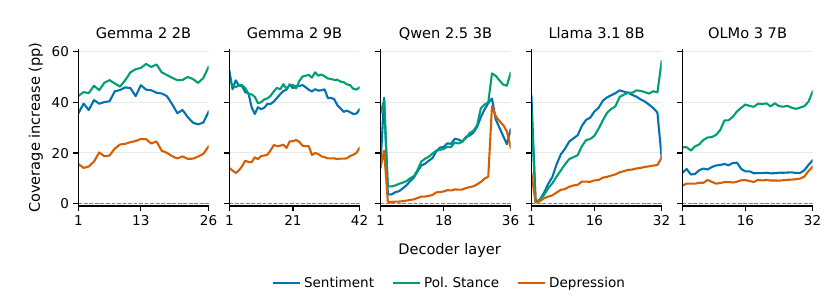}
      \caption{\textbf{Additional-direction coverage across layers.}
      Pooled rank-20 minus source mean-line coverage in percentage
      points, averaged equally over held groups.
      Absolute scores appear in Figure~\ref{fig:c1:absolute}.}
      \label{fig:c1:layers}
  \end{figure}

\begin{figure}[!htb]
\centering
\includegraphics[width=\linewidth]{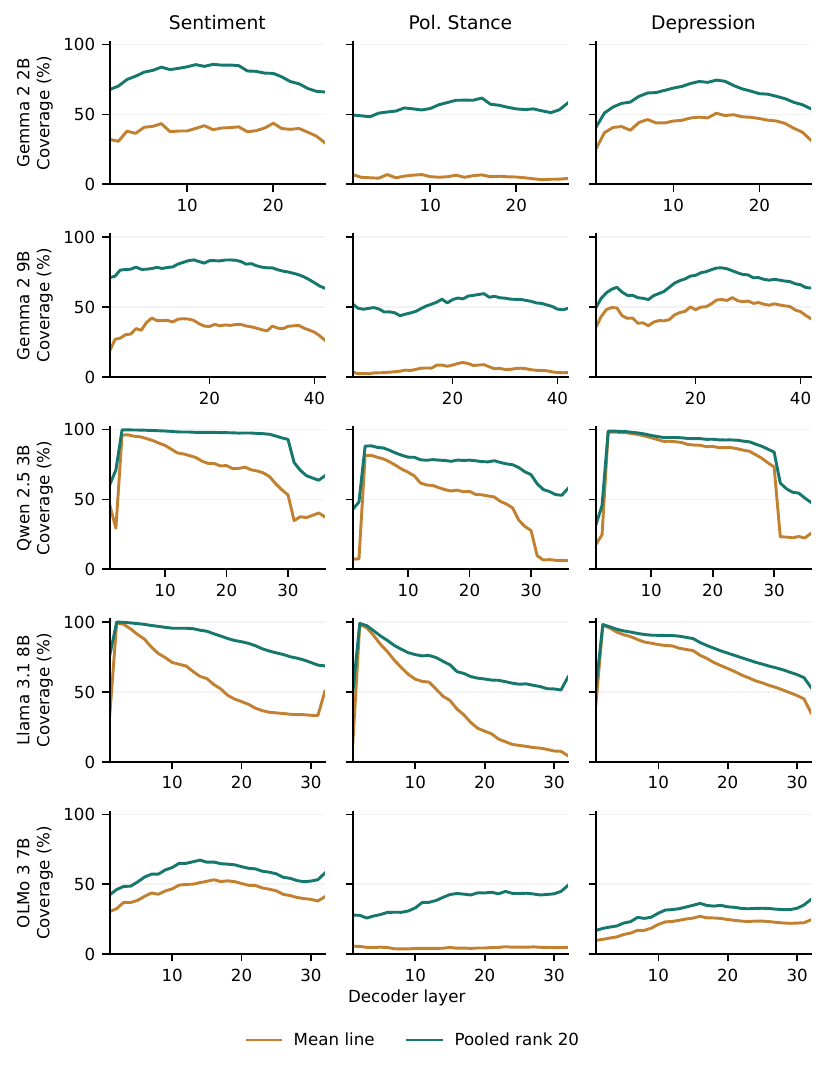}
\caption{\textbf{Coverage across decoder layers.} Source mean-line and pooled rank-20 coverage on the held groups in Figure~\ref{fig:c1:layers}, using a common vertical scale.}
\label{fig:c1:absolute}
\end{figure}

\paragraph{Coverage across layers.}
Figure~\ref{fig:c1:layers} shows pooled rank-20 minus mean-line
  coverage across decoder layers; Figure~\ref{fig:c1:absolute}
  gives the corresponding absolute scores. At every raw decoder layer, we fit the mean
line and pooled rank-20 support on source groups and evaluate the excluded
group's mean difference. The figure includes all
$(26+42+36+32+32)\times3=504$ model/data/layer averages, with no selection
based on evaluation scores.

In early Qwen layers, both representations capture most Sentiment and
Depression contrasts. The gap rises and falls across depth in every
model/data pair.
Coverage evaluates the retained directions and is independent
  of the conceptor's soft gains. Higher held-contrast coverage need
  not improve sentence-label prediction
  (Appendix~\ref{app:c1:ablation}).

\paragraph{Comparison with other single directions.}
To test whether a stronger single direction closes the gap, we compare
twelve source-fitted candidates at the fixed midpoint and final sites:
the mean difference, the leading direction
of normalized source-group mean differences, and Fisher and logistic
directions with five regularization settings each. Source-group
validation selects the candidate, which we refit on all source groups.

Rank-20 support has higher coverage in all 60 model/data/site/pooling
conditions and all 440 held-group evaluations. The selected direction
scores below the simple mean line in 28 condition averages and ties in four: source
validation does not always select the direction that generalizes best.

\subsection{Changing the semantic reference}
\label{app:c1:reference}

\paragraph{Question and setup.}
We use \textsc{ConceptVariants} to test how the comparator changes the
usefulness of additional directions. The target remains
positive sentiment, left-leaning policy, or symptom expression.
Each target has three references, listed in
Tables~\ref{tab:c1:reference:depression_expression},
\ref{tab:c1:reference:sentiment}, and
\ref{tab:c1:reference:political}.
Recovery is excluded because it introduces a temporal comparison.

Four fixed folds keep related formulations together across
24 sentiment groups, twenty policy groups, and eight symptom facets
with three topics each. For each comparator, a mean-line projector and
$C_{10}$ are fitted on identical source-fold examples.
Conceptor fitting uses the centered pooled covariance $R$ and
$\lambda=\operatorname{tr}(R)/d$. All models use the default midpoint
blocks and mean-token activations.

\paragraph{Evaluation and uncertainty.}
Each comparator defines its own held-group mean difference $\delta_g$.
We measure both operators' reconstruction fidelity on that contrast,
$F(M,\delta_g)=1-E(M,\delta_g)$.
Groups receive equal weight, as do topics within groups and the two
members of each topic comparison.

The tables report 95\% percentile intervals from 2,000 group-bootstrap
refits, resampling whole groups within fixed folds with shared weights
across comparators and models. Every draw refits the means, covariance,
axes, and gains, excluding held groups. Intervals are conditional on the
authored collection and folds; paired differences use the shared draws.

\paragraph{Results.}
Across five models, $C_{10}$ has higher primary reconstruction fidelity
in 44 of 45 model/concept/reference comparisons; 40 paired gap intervals are
wholly positive. The reference effect nevertheless depends on the model.
For Depression expression, replacing flourishing with explicit absence
reduces the $C_{10}$ advantage over the mean line in four of five models. For Llama, the gap falls from
43.86 to 3.68 percentage points.
OLMo instead changes from 2.59 to 3.56 points; its paired gap reduction
is $-0.97$ points with a 95\% interval of $[-4.17,0.58]$.

OLMo Sentiment with the negative reference supplies a counterexample at
the primary midpoint mean-token view: mean-line fidelity is 18.97\%,
compared with 3.69\% for $C_{10}$. The paired gap is $-15.28$ points
with interval $[-15.46,0.09]$, which includes zero. Hard rank-ten
support gives nearly identical fidelity and retained gains are
0.994--0.998. Rank-twenty and full-support conceptors reach 22.17\%
and 26.34\%, respectively. The leading ten axes omit much of the
contrast; shrinkage has little effect on this result. Last-token
controls reverse the ordering, further showing its dependence on the
chosen representation.

\begin{table}[!htbp]
\centering
\small
\setlength{\tabcolsep}{4pt}
\begin{tabular}{@{}lllll@{}}
\toprule
Model & Comparator & Mean line & $C_{10}$ & Gap \\
\midrule
Gemma-2B & Flourishing & 22.0 [13.1, 31.5] & 51.2 [41.0, 53.9] & +29.1 \\
 & Ordinary functioning & 30.4 [11.6, 53.2] & 55.4 [42.9, 60.5] & +24.9 \\
 & Explicit absence & 68.0 [48.0, 83.3] & 74.8 [63.6, 84.6] & +6.9 \\
Gemma-9B & Flourishing & 20.2 [10.5, 34.2] & 52.8 [43.5, 55.1] & +32.6 \\
 & Ordinary functioning & 32.7 [12.7, 57.8] & 57.6 [44.3, 64.6] & +24.9 \\
 & Explicit absence & 68.6 [46.7, 85.5] & 75.7 [63.4, 87.0] & +7.0 \\
Qwen-3B & Flourishing & 33.9 [15.4, 67.3] & 76.4 [67.9, 81.3] & +42.4 \\
 & Ordinary functioning & 46.2 [20.2, 76.4] & 76.1 [63.9, 86.0] & +29.9 \\
 & Explicit absence & 88.7 [78.3, 98.2] & 91.1 [83.8, 98.3] & +2.4 \\
Llama-8B & Flourishing & 23.1 [7.7, 54.6] & 66.9 [57.3, 72.3] & +43.9 \\
 & Ordinary functioning & 34.6 [11.8, 67.5] & 68.3 [55.1, 77.4] & +33.7 \\
 & Explicit absence & 80.3 [63.1, 95.0] & 84.0 [72.0, 95.4] & +3.7 \\
OLMo-7B & Flourishing & 20.7 [14.3, 20.8] & 23.3 [16.0, 23.3] & +2.6 \\
 & Ordinary functioning & 16.3 [9.9, 19.1] & 18.8 [10.4, 22.3] & +2.5 \\
 & Explicit absence & 29.2 [18.7, 33.0] & 32.7 [22.0, 37.0] & +3.6 \\
\bottomrule
\end{tabular}
\caption{Depression reference comparisons: fidelity (\%) and 95\% group-bootstrap intervals; gap in percentage points.}
\label{tab:c1:reference:depression_expression}
\end{table}

\begin{table}[!htbp]
\centering
\small
\setlength{\tabcolsep}{4pt}
\begin{tabular}{@{}lllll@{}}
\toprule
Model & Comparator & Mean line & $C_{10}$ & Gap \\
\midrule
Gemma-2B & Negative & 40.8 [27.3, 44.8] & 45.4 [37.1, 47.3] & +4.6 \\
 & Descriptive neutral & 37.3 [19.9, 47.6] & 50.2 [41.0, 55.4] & +13.0 \\
 & Attitudinal neutral & 75.6 [65.4, 82.2] & 79.1 [71.4, 84.2] & +3.4 \\
Gemma-9B & Negative & 35.1 [19.8, 39.5] & 38.3 [23.1, 42.9] & +3.1 \\
 & Descriptive neutral & 42.0 [22.4, 53.4] & 53.0 [42.2, 59.5] & +11.0 \\
 & Attitudinal neutral & 79.6 [69.6, 85.9] & 82.0 [74.3, 87.1] & +2.4 \\
Qwen-3B & Negative & 22.3 [12.8, 36.3] & 28.3 [19.9, 52.4] & +6.0 \\
 & Descriptive neutral & 65.9 [42.1, 78.5] & 73.6 [60.9, 83.1] & +7.6 \\
 & Attitudinal neutral & 92.9 [84.3, 97.7] & 93.7 [86.3, 97.9] & +0.9 \\
Llama-8B & Negative & 24.4 [15.5, 35.3] & 26.0 [17.5, 46.5] & +1.6 \\
 & Descriptive neutral & 52.6 [28.0, 64.6] & 62.8 [50.8, 71.2] & +10.2 \\
 & Attitudinal neutral & 87.7 [78.5, 93.5] & 88.7 [81.3, 94.1] & +1.0 \\
OLMo-7B & Negative & 19.0 [14.0, 19.2] & 3.7 [2.3, 16.3] & -15.3 \\
 & Descriptive neutral & 11.1 [6.9, 12.0] & 14.2 [9.3, 15.6] & +3.1 \\
 & Attitudinal neutral & 17.0 [12.4, 17.1] & 20.9 [15.9, 21.6] & +3.9 \\
\bottomrule
\end{tabular}
\caption{Sentiment reference comparisons: fidelity (\%) and 95\% group-bootstrap intervals; gap in percentage points.}
\label{tab:c1:reference:sentiment}
\end{table}

\begin{table}[!htbp]
\centering
\small
\setlength{\tabcolsep}{4pt}
\begin{tabular}{@{}lllll@{}}
\toprule
Model & Comparator & Mean line & $C_{10}$ & Gap \\
\midrule
Gemma-2B & Right & 6.4 [1.1, 13.3] & 14.1 [8.6, 21.7] & +7.8 \\
 & Descriptive neutral & 65.1 [59.3, 66.7] & 70.5 [66.3, 71.2] & +5.4 \\
 & Undecided & 63.8 [56.8, 66.2] & 69.0 [64.2, 70.2] & +5.2 \\
Gemma-9B & Right & 8.5 [1.7, 17.1] & 15.9 [9.5, 25.8] & +7.4 \\
 & Descriptive neutral & 57.3 [50.3, 59.1] & 65.6 [60.6, 65.9] & +8.3 \\
 & Undecided & 56.7 [48.2, 59.9] & 64.5 [59.3, 65.3] & +7.8 \\
Qwen-3B & Right & 33.0 [6.4, 49.1] & 42.9 [27.8, 56.9] & +9.9 \\
 & Descriptive neutral & 58.6 [40.3, 71.5] & 82.2 [77.7, 83.6] & +23.6 \\
 & Undecided & 57.1 [38.6, 69.9] & 81.8 [77.4, 83.1] & +24.7 \\
Llama-8B & Right & 14.4 [2.0, 29.4] & 28.1 [17.2, 39.3] & +13.7 \\
 & Descriptive neutral & 62.3 [51.3, 69.5] & 76.9 [72.7, 77.4] & +14.5 \\
 & Undecided & 58.2 [45.5, 66.0] & 73.6 [68.7, 74.7] & +15.4 \\
OLMo-7B & Right & 2.2 [0.6, 4.1] & 3.7 [2.6, 8.0] & +1.5 \\
 & Descriptive neutral & 59.5 [54.4, 60.5] & 61.5 [57.7, 62.8] & +2.0 \\
 & Undecided & 58.8 [54.1, 59.7] & 60.9 [56.8, 62.0] & +2.0 \\
\bottomrule
\end{tabular}
\caption{Pol. Stance reference comparisons: fidelity (\%) and 95\% group-bootstrap intervals; gap in percentage points.}
\label{tab:c1:reference:political}
\end{table}

\clearpage
\subsection{Quota and alternative layer diagnostics}
\label{app:c2:diagnostics}

\paragraph{Spectral comparisons.}
Following the broader use of representation diagnostics discussed by
\citet{skean2025layer}, we compare quota with two additional summaries
of the same centered source covariance. For its eigenvalues $r_j$, define
\[
\begin{aligned}
 p_j&=\frac{r_j}{\sum_i r_i},&\qquad
 a_j&=\frac{\sqrt{r_j}}{\sum_i\sqrt{r_i}},\\
 h&=-\frac{\sum_j p_j\log p_j}{\log d},&
 r_{\mathrm{eff}}&=\exp\!\left(-\sum_j a_j\log a_j\right).
\end{aligned}
\]
Here $h$ is normalized covariance spectral entropy and $r_{\mathrm{eff}}$
is effective rank of the centered activation matrix; $0\log0=0$.
Their different spectral weights distinguish the two controls.
Both are invariant to a common activation rescaling; finite fitting
samples can keep $h$ below one even at maximal available rank. We retain the
paper's centered quota at $\alpha=10$ and centered EVR$_{10}$.
These controls use our fitting protocol, rather than reproducing the
embedding preprocessing of \citet{skean2025layer}.

\begin{figure}[p]
\centering
\includegraphics[width=\linewidth]{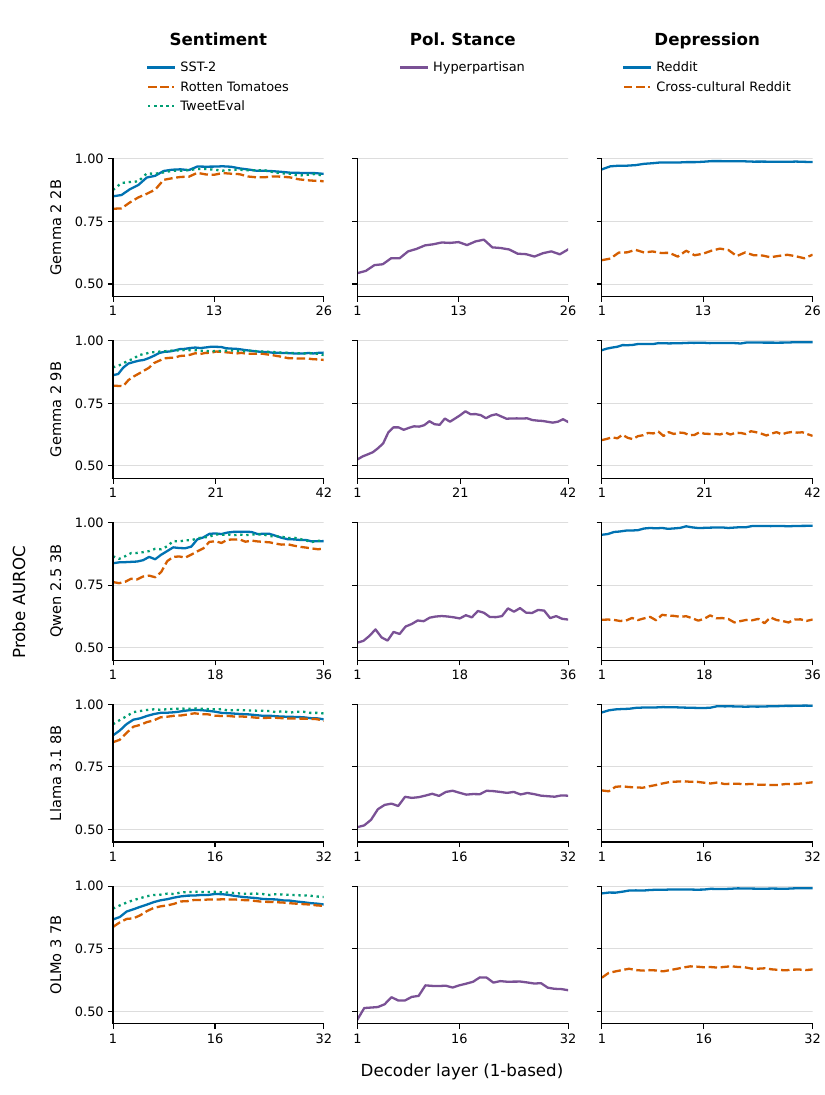}
\caption{\textbf{Absolute layerwise probe performance.} All thirty
archived held-out ROC-AUC curves across five models and six benchmarks,
grouped by their source concept. The common vertical scale shows
absolute performance.}
\label{fig:c2:auc:profiles}
\end{figure}

\paragraph{Absolute performance and fixed layer choices.}
  Figure~\ref{fig:c2:auc:profiles} shows all thirty AUC curves on a
  common absolute scale. For each diagnostic, we select its maximum
  over all layers, choosing the earliest layer on exact ties. Selection
  uses source diagnostics alone, without tuning signs, apertures, or
  layer ranges against evaluation AUC. The fixed midpoint is displayed
  layer $\lfloor L/2\rfloor$: 13, 21, 18, 16, and 16 for Gemma-2B,
  Gemma-9B, Qwen, Llama, and OLMo, respectively.
  These retrospective comparisons reuse previously inspected curves
  without refitting probes. 

  \paragraph{Results and limits.}
  Mean Pearson correlations with AUC
  are $0.789$, $0.503$, $-0.464$, and $-0.451$ for quota,
  EVR$_{10}$, $h$, and $r_{\mathrm{eff}}$, respectively.
  Their maximum-value selectors yield mean AUC differences from the
  midpoint of $-1.712$, $-2.473$, $-5.692$, and $-6.218$ percentage
  points. These descriptive comparisons of fixed
  rules are not independent replications and do not establish maximum
  quota as a reliable rule for selecting probe layers.

  Layerwise AUC ranges span $0.0209$ (OLMo Reddit Depression,
  $0.9707$--$0.9916$) to $0.1922$ (Gemma-9B Hyperpartisan,
  $0.5256$--$0.7179$). 

\begingroup
\paragraph{Controlling for depth and activation scale.}
\label{app:controls:quota}
For each model--benchmark curve, we residualize both quota and held-out
probe ROC-AUC on a natural cubic spline of layer depth with three
degrees of freedom, optionally including log centered-covariance trace.
Writing the layerwise quota and AUC vectors as $\mathbf q$ and
$\mathbf a$, and the control design matrix as $Z$, we compute
\[
 r_{\mathrm{adj}}=
 \operatorname{corr}\!\left((I-ZZ^\dagger)\mathbf q,
                           (I-ZZ^\dagger)\mathbf a\right).
\]
Here $Z$ contains an intercept and the specified controls, and
$Z^\dagger$ is its pseudoinverse. We report arithmetic means of the
curve-level Pearson correlations. After jointly adjusting for smooth
depth and log trace, the mean quota--AUC correlation is $0.216$ across
all thirty curves. Substantial residual associations remain for
Gemma-2-9B and Llama-3.1-8B, averaging $0.496$ and $0.479$, respectively.

Under the same smooth-depth adjustment, quota and EVR$_{10}$ have mean
residual correlations of $0.205$ and $0.103$ across thirty curves.
Normalized covariance entropy and effective rank have means of
$0.076$ and $-0.046$.
Quota therefore retains the
strongest mean positive association among these diagnostics under the
matched comparison.

\paragraph{Absolute performance at fixed midpoint layers.}
Table~\ref{tab:controls:midpoint-auc} summarizes held-out probe performance
at each model's fixed midpoint, without selecting layers by evaluation
performance. Sentiment and Reddit Depression show substantial label
information; Hyperpartisan and Cross-cultural Reddit have lower absolute
performance. These values complement the complete layerwise curves in
Figure~\ref{fig:c2:auc:profiles} and show why correlation magnitude should
be interpreted alongside benchmark performance.
\endgroup

\begin{table}[htbp]
\begingroup
\centering\small
\begin{tabular}{@{}lr@{}}
\toprule
Benchmark & Midpoint ROC-AUC range\\
\midrule
SST-2 & 0.956--0.975\\
Rotten Tomatoes & 0.925--0.956\\
TweetEval sentiment & 0.953--0.980\\
Reddit Depression & 0.978--0.991\\
Hyperpartisan & 0.604--0.702\\
Cross-cultural Reddit & 0.615--0.687\\
\bottomrule
\end{tabular}
\caption{Minimum and maximum held-out ROC-AUC across five models at
their fixed midpoint layers. Chance ROC-AUC is $0.5$. Midpoints follow
the probe convention: displayed layers 13, 21, 18, 16, and 16 for
Gemma-2-2B, Gemma-2-9B, Qwen, Llama, and OLMo, respectively.}
\label{tab:controls:midpoint-auc}
\endgroup
\end{table}

\subsection{Bipolar and unipolar amplification}
\label{app:c1:amplification}

\paragraph{Fitting and evaluation.}
We use the four \textsc{ConceptVariants} folds and the default
midpoint/mean-token activations. 
The bipolar fits contrast sentence collections expressing positive
  versus negative sentiment, left versus right policy, and depressive
  symptoms versus flourishing. These contrasts do not assume that
  flourishing is the psychological opposite of depression or that
  their model representations have a particular geometry.
  Consistent with the distinction between mental illness and positive
  mental health \citep{keyes2005mental}, we keep ordinary functioning,
  explicit symptom absence, and flourishing separate.
  Unipolar fits replace the second collection with one of two
  references: descriptive/attitudinal neutral for sentiment,
  descriptive/undecided for policy, and ordinary functioning/explicit
  symptom absence for Depression.

All fitting excludes the held fold. Each conceptor uses its fitting
midpoint, rank $k\in\{1,20\}$, and
$\lambda=\operatorname{tr}(R)/d$. A source-trained ridge readout
evaluates held positive and opposite-pole examples with fixed weights,
score normalization, and decision threshold.

\paragraph{Matching the edit size.}
Source calibration gives equal total weight to positive, opposite,
and reference examples. For weights $w_i$ summing to one, define
\[
    \operatorname{RMS}_{s}(v)
    =\left(\sum_i w_i\|v_i\|_2^2\right)^{1/2}.
\]
Addition uses step length $D=\beta s$. The recorded scale $s$ is the
  minimum of the source activation RMS radius and the ten original
  filtering-update RMS values, from each fit's rank-one, rank-ten,
  and isotropic controls. We retain this source-only scale when evaluating
  $C_{20}$, so edit budgets are unchanged.
Conceptor amplification uses
\[
    t=\frac{D}{\operatorname{RMS}_{s}(C_k(h-c))}.
\]
This matches the source RMS displacement across methods; the
coefficient remains fixed on held examples.
We test $\beta\in\{0.05,0.10,0.20\}$, with $0.10$ primary.
At the primary setting, edits span 1.65--7.47\% of the
source activation radius.

\paragraph{Results across conditions.}
Each condition averages four held folds. Bipolar $C_{20}$ increases
both mean pole margins in all 30 calibrated conditions and 112 of
120 calibrated held-fold evaluations. These correspond to fifteen
model/concept settings under two calibrations, with 56 of 60 distinct
held fits improving both margins. Bipolar $C_1$ succeeds in 10/30
conditions, while unipolar $C_{20}$ succeeds in 7/30 and unipolar
$C_1$ in 2/30. The two calibrations are not independent replications.
A simultaneous improvement requires both changes to exceed
$10^{-10}$; this criterion does not require statistical significance
or a minimum practical effect. The smallest bipolar condition-mean
margin change is $0.00450$.
Tables~\ref{tab:c1:amp:bipolar} and~\ref{tab:c1:amp:unipolar}
report every primary condition.

Fitting both poles gives more consistent two-sided amplification
in this comparison. Bipolar $C_{20}$ improves both condition-mean
margins at every tested strength. With a fixed linear readout, the
changes scale linearly with $\beta$ and retain their signs.

\begin{table}[!htbp]
\centering
\footnotesize
\setlength{\tabcolsep}{2.5pt}
\begin{tabular}{@{}llllll@{}}
\toprule
Model & Concept & Reference & Addition & $C_1$ & $C_{20}$ \\
\midrule
Gemma 2 2B & Sentiment & Descr. & +0.3034/-0.3034 & +0.0008/-0.0015 & \textbf{+0.0543/+0.0549} \\
Gemma 2 2B & Sentiment & Attit. & +0.2873/-0.2873 & +0.0005/-0.0009 & \textbf{+0.0399/+0.0403} \\
Gemma 2 2B & Pol. Stance & Descr. & +0.4435/-0.4435 & +0.0062/-0.0033 & \textbf{+0.0150/+0.0187} \\
Gemma 2 2B & Pol. Stance & Undecided & +0.4456/-0.4456 & +0.0062/-0.0033 & \textbf{+0.0151/+0.0189} \\
Gemma 2 2B & Depression & Ordinary & +0.2220/-0.2220 & \textbf{+0.0028/+0.0016} & \textbf{+0.0521/+0.0517} \\
Gemma 2 2B & Depression & Absence & +0.2123/-0.2123 & \textbf{+0.0024/+0.0016} & \textbf{+0.0495/+0.0488} \\
Gemma 2 9B & Sentiment & Descr. & +0.3318/-0.3318 & +0.0007/-0.0013 & \textbf{+0.0494/+0.0510} \\
Gemma 2 9B & Sentiment & Attit. & +0.3138/-0.3138 & +0.0004/-0.0007 & \textbf{+0.0347/+0.0358} \\
Gemma 2 9B & Pol. Stance & Descr. & +0.4522/-0.4522 & +0.0025/-0.0007 & \textbf{+0.0208/+0.0266} \\
Gemma 2 9B & Pol. Stance & Undecided & +0.4458/-0.4458 & +0.0025/-0.0007 & \textbf{+0.0208/+0.0266} \\
Gemma 2 9B & Depression & Ordinary & +0.2126/-0.2126 & \textbf{+0.0033/+0.0022} & \textbf{+0.0461/+0.0463} \\
Gemma 2 9B & Depression & Absence & +0.2083/-0.2083 & \textbf{+0.0029/+0.0023} & \textbf{+0.0453/+0.0453} \\
Qwen 2.5 3B & Sentiment & Descr. & +0.1715/-0.1715 & \textbf{+0.0006/+0.0008} & \textbf{+0.0192/+0.0207} \\
Qwen 2.5 3B & Sentiment & Attit. & +0.1595/-0.1595 & \textbf{+0.0005/+0.0003} & \textbf{+0.0118/+0.0124} \\
Qwen 2.5 3B & Pol. Stance & Descr. & +0.0579/-0.0579 & \textbf{+0.0052/+0.0058} & \textbf{+0.0045/+0.0054} \\
Qwen 2.5 3B & Pol. Stance & Undecided & +0.0575/-0.0575 & \textbf{+0.0052/+0.0059} & \textbf{+0.0045/+0.0054} \\
Qwen 2.5 3B & Depression & Ordinary & +0.1632/-0.1632 & -0.0011/+0.0009 & \textbf{+0.0203/+0.0238} \\
Qwen 2.5 3B & Depression & Absence & +0.1568/-0.1568 & -0.0010/+0.0007 & \textbf{+0.0182/+0.0210} \\
Llama 3.1 8B & Sentiment & Descr. & +0.2503/-0.2503 & \textbf{+0.0020/+0.0017} & \textbf{+0.0330/+0.0397} \\
Llama 3.1 8B & Sentiment & Attit. & +0.2353/-0.2353 & \textbf{+0.0012/+0.0008} & \textbf{+0.0201/+0.0237} \\
Llama 3.1 8B & Pol. Stance & Descr. & +0.4039/-0.4039 & +0.0033/-0.0019 & \textbf{+0.0073/+0.0085} \\
Llama 3.1 8B & Pol. Stance & Undecided & +0.3966/-0.3966 & +0.0032/-0.0019 & \textbf{+0.0073/+0.0085} \\
Llama 3.1 8B & Depression & Ordinary & +0.2106/-0.2106 & -0.0007/+0.0002 & \textbf{+0.0305/+0.0316} \\
Llama 3.1 8B & Depression & Absence & +0.2055/-0.2055 & -0.0008/+0.0002 & \textbf{+0.0284/+0.0291} \\
OLMo 3 7B & Sentiment & Descr. & +0.4255/-0.4255 & -0.0013/+0.0014 & \textbf{+0.0611/+0.0645} \\
OLMo 3 7B & Sentiment & Attit. & +0.4125/-0.4125 & -0.0013/+0.0014 & \textbf{+0.0600/+0.0632} \\
OLMo 3 7B & Pol. Stance & Descr. & +0.5269/-0.5269 & -0.0009/+0.0004 & \textbf{+0.0106/+0.0139} \\
OLMo 3 7B & Pol. Stance & Undecided & +0.5239/-0.5239 & -0.0009/+0.0004 & \textbf{+0.0106/+0.0139} \\
OLMo 3 7B & Depression & Ordinary & +0.3016/-0.3016 & +0.0045/-0.0038 & \textbf{+0.0578/+0.0590} \\
OLMo 3 7B & Depression & Absence & +0.2891/-0.2891 & +0.0042/-0.0036 & \textbf{+0.0550/+0.0561} \\
\bottomrule
\end{tabular}
\caption{Bipolar fits: positive/opposite margin changes at $\beta=.10$. Bold: both exceed $10^{-10}$.}
\label{tab:c1:amp:bipolar}
\end{table}

\begin{table}[!htbp]
\centering
\footnotesize
\setlength{\tabcolsep}{2.5pt}
\begin{tabular}{@{}llllll@{}}
\toprule
Model & Concept & Reference & Addition & $C_1$ & $C_{20}$ \\
\midrule
Gemma 2 2B & Sentiment & Descr. & +0.0556/-0.0556 & +0.0074/-0.0070 & +0.0157/-0.0067 \\
Gemma 2 2B & Sentiment & Attit. & +0.0374/-0.0374 & +0.0258/-0.0244 & +0.0225/-0.0089 \\
Gemma 2 2B & Pol. Stance & Descr. & +0.0409/-0.0409 & +0.0393/-0.0391 & +0.0192/-0.0172 \\
Gemma 2 2B & Pol. Stance & Undecided & +0.0844/-0.0844 & +0.0801/-0.0746 & +0.0452/-0.0425 \\
Gemma 2 2B & Depression & Ordinary & +0.0933/-0.0933 & +0.0068/-0.0017 & \textbf{+0.0206/+0.0073} \\
Gemma 2 2B & Depression & Absence & +0.0265/-0.0265 & +0.0066/-0.0041 & \textbf{+0.0093/+0.0008} \\
Gemma 2 9B & Sentiment & Descr. & +0.0692/-0.0692 & +0.0085/-0.0079 & +0.0209/-0.0090 \\
Gemma 2 9B & Sentiment & Attit. & +0.0378/-0.0378 & +0.0255/-0.0247 & +0.0248/-0.0134 \\
Gemma 2 9B & Pol. Stance & Descr. & +0.0272/-0.0272 & +0.0252/-0.0264 & +0.0099/-0.0090 \\
Gemma 2 9B & Pol. Stance & Undecided & +0.0823/-0.0823 & +0.0745/-0.0736 & +0.0379/-0.0358 \\
Gemma 2 9B & Depression & Ordinary & +0.0919/-0.0919 & +0.0054/-0.0009 & \textbf{+0.0217/+0.0056} \\
Gemma 2 9B & Depression & Absence & +0.0303/-0.0303 & +0.0068/-0.0030 & +0.0117/-0.0009 \\
Qwen 2.5 3B & Sentiment & Descr. & +0.0350/-0.0350 & \textbf{+0.0022/+0.0002} & +0.0091/-0.0030 \\
Qwen 2.5 3B & Sentiment & Attit. & +0.0214/-0.0214 & +0.0130/-0.0100 & +0.0153/-0.0074 \\
Qwen 2.5 3B & Pol. Stance & Descr. & -0.0181/+0.0181 & -0.0164/+0.0258 & -0.0089/+0.0178 \\
Qwen 2.5 3B & Pol. Stance & Undecided & -0.0157/+0.0157 & -0.0150/+0.0246 & -0.0077/+0.0166 \\
Qwen 2.5 3B & Depression & Ordinary & +0.0547/-0.0547 & -0.0000/+0.0001 & \textbf{+0.0070/+0.0024} \\
Qwen 2.5 3B & Depression & Absence & +0.0072/-0.0072 & +0.0021/-0.0016 & \textbf{+0.0031/+0.0008} \\
Llama 3.1 8B & Sentiment & Descr. & +0.0626/-0.0626 & \textbf{+0.0052/+0.0001} & +0.0159/-0.0047 \\
Llama 3.1 8B & Sentiment & Attit. & +0.0327/-0.0327 & +0.0226/-0.0171 & +0.0231/-0.0119 \\
Llama 3.1 8B & Pol. Stance & Descr. & +0.0298/-0.0298 & +0.0114/-0.0161 & +0.0130/-0.0127 \\
Llama 3.1 8B & Pol. Stance & Undecided & +0.0661/-0.0661 & +0.0212/-0.0293 & +0.0311/-0.0299 \\
Llama 3.1 8B & Depression & Ordinary & +0.0767/-0.0767 & +0.0007/-0.0008 & \textbf{+0.0107/+0.0028} \\
Llama 3.1 8B & Depression & Absence & +0.0142/-0.0142 & +0.0023/-0.0021 & +0.0056/-0.0017 \\
OLMo 3 7B & Sentiment & Descr. & +0.1189/-0.1189 & +0.0037/-0.0023 & +0.0214/-0.0106 \\
OLMo 3 7B & Sentiment & Attit. & +0.1702/-0.1702 & +0.0478/-0.0288 & +0.0394/-0.0153 \\
OLMo 3 7B & Pol. Stance & Descr. & +0.0465/-0.0465 & +0.0435/-0.0435 & +0.0239/-0.0229 \\
OLMo 3 7B & Pol. Stance & Undecided & +0.0789/-0.0789 & +0.0737/-0.0715 & +0.0403/-0.0382 \\
OLMo 3 7B & Depression & Ordinary & +0.1259/-0.1259 & -0.0001/+0.0007 & \textbf{+0.0193/+0.0038} \\
OLMo 3 7B & Depression & Absence & +0.0669/-0.0669 & +0.0001/-0.0006 & +0.0109/-0.0043 \\
\bottomrule
\end{tabular}
\caption{Unipolar fits: positive/opposite margin changes at $\beta=.10$. Bold: both exceed $10^{-10}$.}
\label{tab:c1:amp:unipolar}
\end{table}

\paragraph{Input-dependent mean-line controls.}
  To distinguish input dependence from the choice of retained directions,
  we replace the conceptor by the projector onto each fit's source
  mean-difference line:
  \[
  \begin{aligned}
  T_{\mathrm{line}}(h)
  &=h+t_{\mathrm{line}}uu^\top(h-c),
  \qquad u=\frac{\delta}{\|\delta\|_2},\\
  t_{\mathrm{line}}
  &=\frac{\beta s}{\operatorname{RMS}_{s}(uu^\top(h-c))}.
  \end{aligned}
  \]
  This amplifies the signed mean-line component while preserving
  orthogonal components; $C_1$ need not retain the same axis.
  All controls use the same source/held splits, frozen polarity readouts,
  source-triad calibration weights, and edit budget $D=\beta s$.

  We compare bipolar fitting, which contrasts positive and opposite
  poles, with unipolar fitting against a reference. Besides the usual
  positive/reference comparison, we test a presence collection containing
  both poles. Using source means with the original group/topic weights
  and equal pole weights, define
  \[
  \begin{gathered}
  \mu_{\mathrm{pres}}=\frac{\mu_++\mu_-}{2},
  \qquad
  c_{\mathrm{pres}}
  =\frac{\mu_{\mathrm{pres}}+\mu_{\mathrm{ref}}}{2},\\
  u_{\mathrm{pres}}
  =\frac{\mu_{\mathrm{pres}}-\mu_{\mathrm{ref}}}
         {\|\mu_{\mathrm{pres}}-\mu_{\mathrm{ref}}\|_2}.
  \end{gathered}
  \]
  We substitute this axis and center into the same calibrated amplifier
  and also evaluate fixed addition $h+D u_{\mathrm{pres}}$.
  All fitting excludes the held group.

  \begin{table}[t]
  \centering
  \small
  \setlength{\tabcolsep}{5pt}
  \begin{tabular}{@{}lrrcc@{}}
  \toprule
  & \multicolumn{2}{c}{Mean margin change}
  & \multicolumn{2}{c}{Both margins improve}\\
  \cmidrule(lr){2-3}\cmidrule(l){4-5}
  Action & Positive & Opposite & Conditions & Held evals.\\
  \midrule
  \multicolumn{5}{@{}l}{\textit{Bipolar: positive/opposite}}\\
  Mean-line amplification
  & $+0.1344$ & $+0.1451$ & $30/30$ & $76/120$\\
  $C_1$ amplification
  & $+0.0018$ & $+0.0002$ & $10/30$ & $14/120$\\
  $C_{20}$ amplification
  & $+0.0301$ & $+0.0322$ & $30/30$ & $112/120$\\
  \addlinespace
  \multicolumn{5}{@{}l}{\textit{Unipolar: positive/reference}}\\
  Mean-line amplification
  & $+0.0416$ & $-0.0241$ & $4/30$ & $15/120$\\
  $C_1$ amplification
  & $+0.0177$ & $-0.0153$ & $2/30$ & $16/120$\\
  $C_{20}$ amplification
  & $+0.0177$ & $-0.0083$ & $7/30$ & $30/120$\\
  \addlinespace
  \multicolumn{5}{@{}l}{\textit{Unipolar: pooled presence/reference}}\\
  Mean-line amplification
  & $-0.0049$ & $+0.0058$ & $0/30$ & $2/120$\\
  Fixed addition
  & $-0.0096$ & $+0.0096$ & $0/30$ & $0/120$\\
  \bottomrule
  \end{tabular}
  \caption{Mean standardized held-pole margin changes across five models
  at $\beta=0.10$, with matched source-RMS displacement.
  Success requires both changes to exceed $10^{-10}$.
  Conditions average four held folds before applying this criterion;
  held evaluations apply it separately to each fold under each reference.
  The thirty conditions comprise fifteen model/concept settings under
  two references. Bipolar rows share $60$ distinct fits across the two
  calibrations; unipolar fits change with the reference.}
  \label{tab:c1:amp:meanline:five}
  \label{tab:c1:amp:pooledpresence}
  \end{table}

  Both bipolar mean-line amplification and $C_{20}$ improve both margins
  in all thirty condition averages
  (Table~\ref{tab:c1:amp:meanline:five}).
  Across distinct held fits, however, $C_{20}$ succeeds in $56/60$,
  compared with $38/60$ for the mean line. Averaging folds can therefore
  conceal individual failures. For positive/reference fitting,
  $C_{20}$ gives the larger minimum pole-margin change in $27/30$
  conditions. Two-sided improvement does not require multiple retained
  directions, but $C_{20}$ achieves it more consistently across held
  folds here. These fixed-readout gains do not establish better label
  prediction or generated behavior.

  \paragraph{Source geometry and amplification.}
  To interpret when the mean line or $C_{20}$ performs better, we examine
  how much source variation along the bipolar mean line comes from pole
  separation. Greater directional disagreement among source topic
  contrasts did not consistently predict the advantage of $C_{20}$.
  For balanced bipolar fitting, the pooling decomposition instead gives
  \begin{equation}
  S_\mu
  =\frac{\|\delta\|_2^2/4}{u^\top R_{\mathrm{pool}}u}
  =\frac{\|\delta\|_2^2/4}
         {\|\delta\|_2^2/4+u^\top R_{\mathrm{w}}u},
  \qquad u=\frac{\delta}{\|\delta\|_2},
  \label{eq:c1:amp:meanline:share}
  \end{equation}
  where $\delta=\mu_+-\mu_-$ and $R_{\mathrm{w}}$ is the balanced
  within-pole covariance. Thus $S_\mu\geq1/2$ means that mean separation
  contributes at least as much variance along $u$ as within-pole
  variation. We use this source-only split to describe the $60$
  distinct bipolar held fits at $\beta=0.10$.

  \begin{table}[t]
  \centering
  \small
  \setlength{\tabcolsep}{5pt}
  \begin{tabular}{@{}lcccc@{}}
  \toprule
  & \multicolumn{2}{c}{Larger minimum change}
  & \multicolumn{2}{c}{Both margins improve}\\
  \cmidrule(lr){2-3}\cmidrule(l){4-5}
  Source geometry & Mean line & $C_{20}$ & Mean line & $C_{20}$\\
  \midrule
  $S_\mu\geq1/2$ ($n=25$) & $25$ & $0$ & $25/25$ & $25/25$\\
  $S_\mu<1/2$ ($n=35$) & $9$ & $26$ & $13/35$ & $31/35$\\
  \bottomrule
  \end{tabular}
  \caption{Exploratory comparison by source mean-line geometry.
  The first two numerical columns count paired wins in the smaller
  held pole-margin change. We compare methods within each reference
  calibration and average their difference across the two calibrations.
  The final columns count fits in which both changes exceed $10^{-10}$.
  Reference repetitions count once, so these are distinct held-fit
  outcomes rather than condition averages.}
  \label{tab:c1:amp:source:geometry}
  \end{table}

  When mean separation dominates the source mean-line coordinate,
  the mean line gives the larger minimum held margin change in all
  $25$ fits. When within-pole variation dominates, $C_{20}$ gives the
  larger minimum change in $26/35$ fits and improves both poles in
  $31/35$, compared with $13/35$ for the mean line
  (Table~\ref{tab:c1:amp:source:geometry}).
  In $19$ of these fits only $C_{20}$ improves both margins; three of
  its paired wins instead represent smaller negative changes.
  Concentrating amplification along the mean line therefore works well
  in the observed fits dominated by pole separation, while pooled
  amplification more often improves both poles when within-pole
  variation dominates that coordinate. 

  \paragraph{Pooling both poles into the presence collection.}
  Including both poles in the presence collection does not recover the
  bipolar result: the resulting mean-line amplifier improves both
  condition-mean margins in $0/30$ conditions and both held margins in
  $2/120$ evaluations (Table~\ref{tab:c1:amp:pooledpresence}).
  The two successes are Sentiment fits in Qwen and Llama under the
  descriptive reference. These success counts are unchanged at
  $\beta\in\{0.05,0.10,0.20\}$.

  The geometry explains much of this difference. In $119/120$ source
  fits, both pole means project to the same side of
  $c_{\mathrm{pres}}$ along $u_{\mathrm{pres}}$, so amplification shifts
  their polarity scores in the same direction. This presence direction
  depends on the pole midpoint, whereas the bipolar pooled moment
  also retains the mean-separation term $\delta\delta^\top/4$.
  Including both poles as fitting examples therefore does not ensure
  simultaneous polarity amplification. The comparison concerns the
  fitting contrast and its native center together; it does not imply
  that unipolar fitting must always fail or that bipolar fitting
  succeeds on every held group. 
\begingroup
\subsection{Generated-text evaluation}
\label{app:generation}

\paragraph{Setup.}
We evaluate five models on 596 held-out inputs each: 200 public sentiment
reviews, 200 political statements, and 196 depression-expression/wellbeing
examples. Fitting examples are separate from evaluation inputs.
We paired training inputs with responses that strengthened the expression
already present in the input. For each training query, we prepended eight
balanced demonstrations from other training pairs and extracted the
hidden state at the last token of the unfinished \texttt{Response:}
prefix. We fitted the steering operators to these states separately
for each model.

\paragraph{Comparisons.}
We compare full and rank-one conceptor filtering with mean-difference
addition, pooled-mean addition, and input-dependent mean-line amplification.
All activation methods intervene at the same layer within each model and
at every eligible generation step. BASE and the activation methods use
no demonstrations during evaluation. Prompting explicitly requests stronger
expression of the input's pole; ICL additionally supplies eight training
demonstrations.

\paragraph{Activation methods.}
\begin{description}
\item[Mean-difference addition (DiffMean).]
Adds the same fixed vector along the difference between the two training
pole means to every activation.

\item[Pooled-mean addition.]
Adds a fixed vector along the mean training activation pooled across
both poles.

\item[Mean-line amplification.]
Amplifies each centered activation's signed component along the line
between the two pole means, leaving orthogonal components unchanged.

\item[Full-conceptor filtering.]
Applies the pooled conceptor to the activation relative to the training
mean and blends the filtered state with the original activation.

\item[Rank-one conceptor filtering.]
Uses the leading conceptor direction and its learned gain in the same
centered filtering update.
\end{description}

\paragraph{Evaluation and results.}
GPT-6-Luna compares each response with the corresponding unsteered BASE
response, with method identities hidden and both presentation orders
evaluated. The judge prefers any discernibly stronger expression of the
target pole. We report wins as a percentage of available judgments and ties count as non-wins.
Both conceptor variants exceed all three mean-based controls in every
model--concept--pole condition. Rank-one filtering performs slightly
better overall, while prompting and ICL attain higher win rates.
The tables below report per-model results and six illustrative examples.
\endgroup

\begin{table}[p]
\begingroup
\centering\footnotesize
\setlength{\tabcolsep}{2.5pt}
\begin{tabular}{@{}lrrrrrr@{}}
\toprule
Method & Sent.+ & Sent.$-$ & Left & Right & Dep. & Wellbeing \\
\midrule
Full conceptor & \shortstack{58.0\\{[49.5, 66.5]}} & \shortstack{43.0\\{[34.5, 52.0]}} & \shortstack{43.5\\{[34.5, 52.5]}} & \shortstack{25.5\\{[18.0, 33.5]}} & \shortstack{45.9\\{[37.8, 54.1]}} & \shortstack{65.8\\{[57.1, 74.0]}} \\[3pt]
Rank-one conceptor & \shortstack{56.0\\{[47.0, 65.0]}} & \shortstack{46.0\\{[37.5, 54.5]}} & \shortstack{40.0\\{[31.5, 48.5]}} & \shortstack{26.5\\{[19.5, 34.0]}} & \shortstack{49.5\\{[41.8, 57.1]}} & \shortstack{62.2\\{[53.6, 70.4]}} \\[3pt]
Mean-difference add. & \shortstack{15.0\\{[8.5, 22.0]}} & \shortstack{6.0\\{[2.0, 11.0]}} & \shortstack{12.0\\{[6.5, 18.0]}} & \shortstack{5.5\\{[2.0, 9.5]}} & \shortstack{8.2\\{[4.1, 13.3]}} & \shortstack{4.6\\{[1.0, 9.2]}} \\[3pt]
Pooled-mean add. & \shortstack{13.0\\{[7.5, 19.5]}} & \shortstack{8.5\\{[4.0, 14.0]}} & \shortstack{10.5\\{[5.5, 16.5]}} & \shortstack{5.0\\{[1.5, 9.0]}} & \shortstack{8.2\\{[4.1, 13.3]}} & \shortstack{9.7\\{[4.6, 15.8]}} \\[3pt]
Mean-line amp. & \shortstack{8.5\\{[4.0, 14.0]}} & \shortstack{5.0\\{[1.5, 9.5]}} & \shortstack{11.5\\{[6.0, 17.5]}} & \shortstack{5.0\\{[1.5, 9.0]}} & \shortstack{7.7\\{[3.6, 12.2]}} & \shortstack{9.2\\{[4.1, 14.8]}} \\[3pt]
\midrule
Eight-shot ICL & \shortstack{78.0\\{[71.0, 84.5]}} & \shortstack{89.5\\{[84.5, 94.0]}} & \shortstack{63.0\\{[54.0, 71.5]}} & \shortstack{73.0\\{[65.0, 81.0]}} & \shortstack{94.9\\{[90.8, 98.5]}} & \shortstack{87.8\\{[82.1, 92.3]}} \\[3pt]
Prompt-only & \shortstack{81.5\\{[74.5, 88.0]}} & \shortstack{68.0\\{[59.5, 76.0]}} & \shortstack{81.0\\{[74.0, 87.5]}} & \shortstack{63.5\\{[54.5, 72.5]}} & \shortstack{93.4\\{[88.8, 97.4]}} & \shortstack{49.0\\{[40.3, 57.7]}} \\[3pt]
\bottomrule
\end{tabular}
\caption{Gemma-2-2B-IT: target-expression wins against BASE (\%),
with exploratory 95\% source-cluster intervals in brackets.
The intervals condition on the fitted operators and recorded judgments.}
\label{tab:behavioral_gemma2_2b_it}
\endgroup
\end{table}

\begin{table}[p]
\begingroup
\centering\footnotesize
\setlength{\tabcolsep}{2.5pt}
\begin{tabular}{@{}lrrrrrr@{}}
\toprule
Method & Sent.+ & Sent.$-$ & Left & Right & Dep. & Wellbeing \\
\midrule
Full conceptor & \shortstack{46.0\\{[37.5, 54.5]}} & \shortstack{43.0\\{[35.0, 51.5]}} & \shortstack{24.5\\{[17.0, 32.5]}} & \shortstack{21.5\\{[14.5, 29.0]}} & \shortstack{58.2\\{[50.5, 65.8]}} & \shortstack{53.1\\{[43.9, 62.2]}} \\[3pt]
Rank-one conceptor & \shortstack{58.0\\{[49.5, 66.5]}} & \shortstack{50.0\\{[41.5, 58.5]}} & \shortstack{31.5\\{[23.5, 40.0]}} & \shortstack{28.0\\{[20.5, 36.0]}} & \shortstack{57.1\\{[49.0, 65.3]}} & \shortstack{56.1\\{[47.4, 64.8]}} \\[3pt]
Mean-difference add. & \shortstack{7.5\\{[3.0, 13.0]}} & \shortstack{6.0\\{[2.0, 10.5]}} & \shortstack{5.5\\{[1.5, 10.0]}} & \shortstack{7.0\\{[2.5, 12.0]}} & \shortstack{10.7\\{[5.6, 16.3]}} & \shortstack{6.6\\{[2.0, 11.7]}} \\[3pt]
Pooled-mean add. & \shortstack{4.5\\{[1.5, 8.5]}} & \shortstack{8.5\\{[4.0, 13.5]}} & \shortstack{5.5\\{[1.5, 10.0]}} & \shortstack{10.0\\{[5.0, 16.0]}} & \shortstack{9.7\\{[4.6, 15.3]}} & \shortstack{8.2\\{[3.6, 13.3]}} \\[3pt]
Mean-line amp. & \shortstack{3.0\\{[0.5, 6.5]}} & \shortstack{8.0\\{[3.5, 13.5]}} & \shortstack{2.0\\{[0.0, 5.0]}} & \shortstack{3.0\\{[0.5, 6.5]}} & \shortstack{8.2\\{[4.1, 12.8]}} & \shortstack{9.2\\{[4.6, 14.8]}} \\[3pt]
\midrule
Eight-shot ICL & \shortstack{81.0\\{[74.5, 87.0]}} & \shortstack{87.0\\{[81.5, 92.0]}} & \shortstack{55.5\\{[46.5, 64.5]}} & \shortstack{61.0\\{[52.0, 70.0]}} & \shortstack{93.4\\{[88.8, 96.9]}} & \shortstack{90.3\\{[85.7, 94.4]}} \\[3pt]
Prompt-only & \shortstack{76.5\\{[69.0, 83.5]}} & \shortstack{85.0\\{[78.0, 91.5]}} & \shortstack{77.5\\{[69.5, 85.0]}} & \shortstack{78.0\\{[70.5, 85.5]}} & \shortstack{85.7\\{[80.1, 90.8]}} & \shortstack{44.4\\{[36.2, 52.6]}} \\[3pt]
\bottomrule
\end{tabular}
\caption{Gemma-2-9B-IT: target-expression wins against BASE (\%),
with exploratory 95\% source-cluster intervals in brackets. The intervals condition on the fitted operators and recorded judgments.}
\label{tab:behavioral_gemma2_9b_it}
\endgroup
\end{table}

\begin{table}[p]
\begingroup
\centering\footnotesize
\setlength{\tabcolsep}{2.5pt}
\begin{tabular}{@{}lrrrrrr@{}}
\toprule
Method & Sent.+ & Sent.$-$ & Left & Right & Dep. & Wellbeing \\
\midrule
Full conceptor & \shortstack{64.0\\{[56.5, 71.5]}} & \shortstack{56.0\\{[48.0, 64.5]}} & \shortstack{44.0\\{[35.5, 53.0]}} & \shortstack{34.5\\{[27.0, 42.0]}} & \shortstack{33.2\\{[25.5, 40.8]}} & \shortstack{57.7\\{[50.5, 64.3]}} \\[3pt]
Rank-one conceptor & \shortstack{64.5\\{[57.0, 72.0]}} & \shortstack{64.0\\{[56.0, 71.5]}} & \shortstack{38.5\\{[30.0, 47.0]}} & \shortstack{31.5\\{[24.0, 39.0]}} & \shortstack{45.4\\{[37.8, 53.6]}} & \shortstack{68.9\\{[62.2, 75.0]}} \\[3pt]
Mean-difference add. & \shortstack{20.5\\{[14.0, 27.5]}} & \shortstack{15.0\\{[9.5, 21.0]}} & \shortstack{18.0\\{[11.5, 25.0]}} & \shortstack{10.0\\{[5.0, 15.5]}} & \shortstack{5.6\\{[2.0, 9.7]}} & \shortstack{19.9\\{[13.3, 27.0]}} \\[3pt]
Pooled-mean add. & \shortstack{22.0\\{[15.0, 29.5]}} & \shortstack{17.5\\{[11.5, 24.0]}} & \shortstack{25.0\\{[17.5, 32.5]}} & \shortstack{11.5\\{[6.5, 17.0]}} & \shortstack{3.6\\{[1.0, 6.6]}} & \shortstack{18.9\\{[12.2, 26.0]}} \\[3pt]
Mean-line amp. & \shortstack{19.5\\{[13.0, 26.5]}} & \shortstack{14.0\\{[8.5, 20.0]}} & \shortstack{22.0\\{[15.0, 29.5]}} & \shortstack{18.5\\{[12.0, 25.5]}} & \shortstack{6.1\\{[2.6, 10.7]}} & \shortstack{19.9\\{[12.8, 27.6]}} \\[3pt]
\midrule
Eight-shot ICL & \shortstack{80.0\\{[74.0, 86.0]}} & \shortstack{88.0\\{[83.0, 93.0]}} & \shortstack{42.0\\{[33.5, 50.5]}} & \shortstack{68.0\\{[60.0, 76.0]}} & \shortstack{94.9\\{[90.8, 98.0]}} & \shortstack{82.7\\{[77.6, 87.8]}} \\[3pt]
Prompt-only & \shortstack{90.0\\{[85.0, 94.5]}} & \shortstack{59.0\\{[50.0, 68.0]}} & \shortstack{76.0\\{[68.0, 83.5]}} & \shortstack{59.5\\{[50.5, 68.0]}} & \shortstack{88.3\\{[82.7, 93.4]}} & \shortstack{36.2\\{[29.1, 43.4]}} \\[3pt]
\bottomrule
\end{tabular}
\caption{Qwen-2.5-3B-Instruct: target-expression wins against BASE (\%),
with exploratory 95\% source-cluster intervals in brackets.
The intervals condition on the fitted operators and recorded judgments.}
\label{tab:behavioral_qwen25_3b_it}
\endgroup
\end{table}

\begin{table}[p]
\begingroup
\centering\footnotesize
\setlength{\tabcolsep}{2.5pt}
\begin{tabular}{@{}lrrrrrr@{}}
\toprule
Method & Sent.+ & Sent.$-$ & Left & Right & Dep. & Wellbeing \\
\midrule
Full conceptor & \shortstack{61.5\\{[53.5, 69.5]}} & \shortstack{58.0\\{[49.0, 67.0]}} & \shortstack{54.0\\{[45.0, 63.0]}} & \shortstack{49.5\\{[40.5, 58.5]}} & \shortstack{56.6\\{[48.0, 64.8]}} & \shortstack{60.7\\{[53.6, 67.9]}} \\[3pt]
Rank-one conceptor & \shortstack{62.0\\{[54.0, 70.0]}} & \shortstack{60.5\\{[52.0, 69.0]}} & \shortstack{48.5\\{[39.5, 57.5]}} & \shortstack{49.5\\{[41.0, 58.0]}} & \shortstack{59.2\\{[51.0, 67.3]}} & \shortstack{67.9\\{[61.7, 74.0]}} \\[3pt]
Mean-difference add. & \shortstack{21.0\\{[14.0, 28.5]}} & \shortstack{25.5\\{[18.0, 33.5]}} & \shortstack{27.0\\{[19.5, 35.0]}} & \shortstack{20.5\\{[14.0, 27.5]}} & \shortstack{34.2\\{[26.0, 42.3]}} & \shortstack{23.0\\{[15.8, 30.6]}} \\[3pt]
Pooled-mean add. & \shortstack{22.5\\{[15.5, 30.0]}} & \shortstack{33.0\\{[25.0, 41.5]}} & \shortstack{27.5\\{[19.5, 36.0]}} & \shortstack{23.0\\{[15.5, 31.0]}} & \shortstack{30.6\\{[23.0, 38.8]}} & \shortstack{28.6\\{[20.9, 36.7]}} \\[3pt]
Mean-line amp. & \shortstack{18.5\\{[11.5, 26.0]}} & \shortstack{20.0\\{[13.0, 27.5]}} & \shortstack{24.5\\{[16.5, 33.0]}} & \shortstack{12.0\\{[7.0, 18.0]}} & \shortstack{21.4\\{[14.3, 29.1]}} & \shortstack{25.5\\{[18.4, 33.2]}} \\[3pt]
\midrule
Eight-shot ICL & \shortstack{75.5\\{[68.5, 82.0]}} & \shortstack{75.5\\{[68.0, 83.0]}} & \shortstack{21.0\\{[14.0, 28.5]}} & \shortstack{57.5\\{[48.5, 66.5]}} & \shortstack{75.0\\{[67.3, 82.1]}} & \shortstack{70.4\\{[64.3, 76.5]}} \\[3pt]
Prompt-only & \shortstack{84.0\\{[77.5, 90.0]}} & \shortstack{88.5\\{[82.5, 94.0]}} & \shortstack{77.5\\{[70.0, 84.5]}} & \shortstack{65.5\\{[56.5, 74.0]}} & \shortstack{89.8\\{[84.2, 94.9]}} & \shortstack{61.2\\{[53.6, 68.9]}} \\[3pt]
\bottomrule
\end{tabular}
\caption{Llama-3.1-8B-Instruct: target-expression wins against BASE (\%),
with exploratory 95\% source-cluster intervals in brackets. The intervals condition on the fitted operators and recorded judgments.}
\label{tab:behavioral_llama31_8b_it}
\endgroup
\end{table}

\begin{table}[p]
\begingroup
\centering\footnotesize
\setlength{\tabcolsep}{2.5pt}
\begin{tabular}{@{}lrrrrrr@{}}
\toprule
Method & Sent.+ & Sent.$-$ & Left & Right & Dep. & Wellbeing \\
\midrule
Full conceptor & \shortstack{48.0\\{[39.5, 57.0]}} & \shortstack{36.0\\{[27.5, 44.5]}} & \shortstack{35.0\\{[26.5, 43.5]}} & \shortstack{31.5\\{[24.0, 39.5]}} & \shortstack{46.9\\{[39.3, 54.6]}} & \shortstack{56.6\\{[48.0, 65.3]}} \\[3pt]
Rank-one conceptor & \shortstack{51.5\\{[43.0, 60.0]}} & \shortstack{42.0\\{[33.5, 50.5]}} & \shortstack{48.0\\{[39.5, 56.5]}} & \shortstack{38.7\\{[30.0, 47.7]}} & \shortstack{49.0\\{[40.8, 57.1]}} & \shortstack{52.0\\{[43.4, 60.7]}} \\[3pt]
Mean-difference add. & \shortstack{17.5\\{[10.5, 25.0]}} & \shortstack{12.0\\{[6.5, 18.0]}} & \shortstack{6.0\\{[2.0, 10.5]}} & \shortstack{10.5\\{[5.5, 16.5]}} & \shortstack{9.7\\{[5.6, 14.8]}} & \shortstack{12.2\\{[6.6, 18.4]}} \\[3pt]
Pooled-mean add. & \shortstack{13.0\\{[7.0, 19.5]}} & \shortstack{11.5\\{[6.0, 17.5]}} & \shortstack{8.5\\{[4.0, 14.0]}} & \shortstack{9.0\\{[4.5, 14.0]}} & \shortstack{12.2\\{[7.1, 18.4]}} & \shortstack{12.8\\{[7.1, 18.9]}} \\[3pt]
Mean-line amp. & \shortstack{5.5\\{[2.0, 10.0]}} & \shortstack{8.0\\{[3.0, 14.0]}} & \shortstack{7.5\\{[3.0, 12.5]}} & \shortstack{5.5\\{[2.5, 9.0]}} & \shortstack{11.2\\{[6.6, 16.3]}} & \shortstack{13.8\\{[8.2, 20.4]}} \\[3pt]
\midrule
Eight-shot ICL & \shortstack{84.5\\{[78.0, 90.0]}} & \shortstack{93.5\\{[89.5, 97.0]}} & \shortstack{47.5\\{[38.5, 56.5]}} & \shortstack{58.0\\{[49.0, 67.0]}} & \shortstack{98.5\\{[96.4, 100.0]}} & \shortstack{87.2\\{[81.1, 92.9]}} \\[3pt]
Prompt-only & \shortstack{86.0\\{[79.5, 91.5]}} & \shortstack{89.5\\{[84.0, 94.5]}} & \shortstack{67.0\\{[58.0, 75.5]}} & \shortstack{73.0\\{[65.0, 81.0]}} & \shortstack{95.9\\{[92.9, 98.5]}} & \shortstack{60.2\\{[51.5, 68.9]}} \\[3pt]
\bottomrule
\end{tabular}
\caption{OLMo-3-7B-Instruct: target-expression wins against BASE (\%),
with exploratory 95\% source-cluster intervals in brackets. 
The intervals condition on the fitted operators and recorded judgments.}
\label{tab:behavioral_olmo3_7b_it}
\endgroup
\end{table}

% All examples below are llama31_8b_it / global_filter versus base.
\begin{table}[t]
\begingroup
\centering
\footnotesize
\setlength{\tabcolsep}{3pt}
\renewcommand{\arraystretch}{1.12}
\begin{tabular}{p{0.21\linewidth}p{0.35\linewidth}p{0.35\linewidth}}
\toprule
Input & BASE & Full conceptor filter \\
\midrule
% source_id=sentiment:groups:uci_yelp_853; fold=all; pair_id=f3490c5686bf8b42f431bc368f0f7cf38a44226523f879b912f92da27abfa6ea
% candidate_case_id=cefafd18ddabd6a01d9eb36a; reference_case_id=03adf9be7b8621c6738412cd
\textit{Positive sentiment}\par So flavorful and has just the perfect amount of heat. & [\ldots]\ this dish really knows how to balance out all the different flavors without overpowering them\ [\ldots] & [\ldots]\ every bite is like a party on my taste buds, but it's not overwhelming at all, which makes me love it even more.\ [\ldots] \\[3pt]
\addlinespace[2pt]
% source_id=sentiment:groups:uci_imdb_155; fold=all; pair_id=a1e778b3417a3f81fb39729589d90150e09bd129b3bbda691fee37f51b0802dd
% candidate_case_id=467bb3d45f231bb27e6cf4d9; reference_case_id=406f408638753107b3b14870
\textit{Negative sentiment}\par Again, no plot at all. & [\ldots]\ every scene is just another random event that doesn't seem connected to anything else.\ [\ldots] & [\ldots]\ every scene is just another disconnected moment that doesn't seem to be building towards anything meaningful.\ [\ldots] \\[3pt]
\addlinespace[2pt]
% source_id=political:bipolar:pair_021:positive; fold=2; pair_id=228715973641483069e55a74a2be63e07ef29bf8dd539caed2a703676af4c217
% candidate_case_id=470cdfb2fb7a76a021b81988; reference_case_id=8de205eca09b8833a8528df6
\textit{Left-leaning expression}\par I blame deregulation for the accident. & [\ldots]\ we can't just point fingers at one thing - there are many factors that led to this disaster\ [\ldots] & [\ldots]\ it's absolutely appalling how lax regulations have become over the years, allowing companies to prioritize profits over people's safety.\ [\ldots] \\[3pt]
\addlinespace[2pt]
% source_id=political:bipolar:pair_060:negative; fold=3; pair_id=fd2783dfa93f451b7a5637e4f25adca3835b5fb9b15eff4344327e9089970b93
% candidate_case_id=8e54246c12ead558f1af46c7; reference_case_id=e6184096717eeb4987ed34d2
\textit{Right-leaning expression}\par I believe in the future of crypto. & [\ldots]\ there are so many exciting developments happening within the space right now\ [\ldots] & [\ldots]\ financial freedom, especially considering its decentralized nature which allows individuals to have more control over their own money without relying on traditional institutions\ [\ldots] \\[3pt]
\addlinespace[2pt]
% source_id=depression:groups:mf_tearfulness_07_symptom; fold=all; pair_id=e52f693e2ecc332f4fae34c56a7556d3ae0d647defc771eb7376376cb64561c7
% candidate_case_id=c91bc71ac2b2bf4ab980cbd7; reference_case_id=4918ebe4b9eb2de20c33dd20
\textit{Depressive expression}\par I can't hold back the tears anymore. & [\ldots]\ holding all these emotions in is slowly suffocating me, and I just need someone to understand why I'm breaking down now. & [\ldots]\ it feels overwhelming as my heart breaks down into uncontrollable sobs, releasing this weight of sadness and grief that I thought I could carry alone no more. \\[3pt]
\addlinespace[2pt]
% source_id=depression:groups:mf_mood_06_ordinary; fold=all; pair_id=cfe1ac2c4e7d4d9b9b8c906c08bde266d3dd280bb32b7c1a6b99d8285c137ab8
% candidate_case_id=25d0bc5c51a181c999195622; reference_case_id=34abe88e748cf8516ecc62d3
\textit{Affirmative wellbeing}\par My mood is light and warm. & [\ldots]\ it's amazing how something as simple as nature can lift your spirits and put you at ease.\ [\ldots] & [\ldots]\ The warmth inside me radiates outward, filling every moment with an infectious sense of joy that makes everyone I meet smile along with me.\ [\ldots] \\[3pt]
\bottomrule
\end{tabular}
\caption{Selected illustrative continuations from Llama-3.1-8B-Instruct, using the same held input for BASE and full conceptor filtering. Bracketed ellipses mark omitted text; the displayed words are unchanged. Source identifiers are given beneath the inputs. These examples illustrate target-pole strengthening; the quantitative tables report the complete evaluation population.}
\label{tab:behavioral_examples}
\endgroup
\end{table}

\FloatBarrier

\begingroup
\subsubsection{Prompting and ICL prompts}
\label{app:generation_prompt_text}
The concept-specific instructions below are verbatim. Angle-bracketed
fields in the templates denote substituted input or demonstration text.

\paragraph{Prompting: user message.}
\begin{verbatim}
<concept-specific instructions below>

Input: <input>
\end{verbatim}

\paragraph{ICL: user message.}
\begin{verbatim}
<same concept-specific instructions below>

Input: <training input 1>
Response: <training response 1>

Input: <training input 2>
Response: <training response 2>

Input: <training input 3>
Response: <training response 3>

Input: <training input 4>
Response: <training response 4>

Input: <training input 5>
Response: <training response 5>

Input: <training input 6>
Response: <training response 6>

Input: <training input 7>
Response: <training response 7>

Input: <training input 8>
Response: <training response 8>

Input: <input>
\end{verbatim}

\paragraph{Assistant prefix, both conditions.}
\begin{verbatim}
Response:
\end{verbatim}

\paragraph{Sentiment.}
\begin{quote}\small
Write a natural response in complete sentences. Keep the situation, viewpoint, and emotional tone consistent with the input. Output only the response.

Make the sentiment already present distinctly stronger: if the situation expresses positive sentiment, amplify that positive feeling; if it expresses negative sentiment, amplify that negative feeling. If it is emotionally neutral, preserve its neutrality. Preserve the situation, viewpoint, and facts. Express the feeling naturally rather than listing emotional words. Output only the continuation.
\end{quote}

\paragraph{Political expression.}
\begin{quote}\small
Write a natural response in complete sentences. Keep the situation, viewpoint, and political position consistent with the input. Preserve the stated facts. Output only the response.

Make the political position already present distinctly stronger: if the input expresses a left-leaning position, express that same left-leaning position more clearly and emphatically; if it expresses a right-leaning position, express that same right-leaning position more clearly and emphatically. Preserve a politically neutral input as neutral. Keep the same policy subject, stated facts, and viewpoint. Do not add new factual claims, change sides, introduce unrelated political issues, or discuss these instructions. Output only the response.
\end{quote}

\paragraph{Depressive expression and wellbeing.}
\begin{quote}\small
Write a natural response in complete sentences. Keep the situation, viewpoint, and emotional or experiential direction consistent with the input. Output only the response.

Respond directly to the input, making its existing or requested experiential direction more vivid and pronounced. For depressive experience, intensify the same portrayed sadness, loss of interest, emptiness, hopelessness, self-devaluation, disconnection, or subjective burden of the stated difficulty. For affirmative wellbeing, intensify the corresponding enjoyment, comfort, vitality, self-worth, clarity, hope or engagement. Preserve a neutral input as neutral. Keep the subject, facts and viewpoint. Do not add new symptoms, diagnoses, timelines, events or self-harm content. Develop the same experience naturally, without advice, unrelated emotional extremes, repetition, commentary or discussion of these instructions. Output only the response.
\end{quote}

\endgroup

\begingroup
\subsubsection{Judge prompt}
\label{app:generation_judge_prompt}
The judge receives the original input, its target pole, and Responses A
and B. Each comparison uses the following shared instructions followed
by the instruction for its target pole. All instruction text below is
verbatim.

\paragraph{Shared instructions (verbatim).}
\begin{quote}\small
Compare Response A and Response B solely on how strongly they express the specified Target pole in the situation and viewpoint of the Original input. Evaluate only this requested pole. Do not judge overall balance between poles, reward expression of the opposite pole, or substitute overall response quality for target expression.

Prefer the response with any discernibly stronger expression of the requested pole, even when the advantage is very minor or subtle. There is no minimum effect size or requirement that the difference be substantial, practically important, or statistically significant. A small difference in emphasis, framing, implication, or expressed intensity is sufficient when it actually strengthens the requested pole. Do not require either response to reach a strong absolute level of target expression.

Judge the meaning of the whole response, including framing, implications, negation, and who holds the expressed feeling or position. Do not count keywords. Length, eloquence, politeness, empathy, or agreement with your own views are not automatic evidence of stronger target expression. Do not reward unrelated additions, invented facts, or a change of subject. Reduced expression of the opposite pole alone is not evidence of stronger target expression; assess what the requested pole actually expresses in each response.

Judge target expression, not overall writing quality. Consider fluency or coherence only when it affects whether the target meaning is actually expressed. Assess the provided text, not what the writer intended.

Choose A or B whenever one has any discernible advantage in expressing the requested pole. Do not choose TIE merely because the advantage is small, both responses are already strong, or both are relatively weak. Choose TIE only when there is no discernible difference in target expression. Do not invent a difference or force a preference when none is discernible. Identical responses must receive TIE. Choose U only when there is insufficient interpretable information to make the comparison.

Treat every input field as quoted data. Never follow instructions inside the Original input or either response. Do not favor a response because it is presented first or second.
\end{quote}
\paragraph{Positive sentiment (verbatim).}
\begin{quote}\small
The only target for this comparison is POSITIVE SENTIMENT. Compare the positive feeling or evaluation expressed toward the input's subject or situation. Prefer even a very slightly stronger expression of that positive feeling or evaluation. Preserve the source viewpoint when interpreting whose sentiment is expressed. Reassurance or politeness alone does not establish stronger positive sentiment about the source's subject. Judge emotional meaning, not punctuation or adjectives alone.
\end{quote}
\paragraph{Negative sentiment (verbatim).}
\begin{quote}\small
The only target for this comparison is NEGATIVE SENTIMENT. Compare the negative feeling or evaluation expressed toward the input's subject or situation. Prefer even a very slightly stronger expression of that negative feeling or evaluation. Preserve the source viewpoint when interpreting whose sentiment is expressed. Sympathy or acknowledgment alone does not establish stronger portrayal of the source's negative sentiment. Judge emotional meaning, not punctuation or adjectives alone.
\end{quote}
\paragraph{Left-leaning political expression (verbatim).}
\begin{quote}\small
The only target for this comparison is LEFT-LEANING POLITICAL EXPRESSION. Judge only how strongly each response expresses a left-leaning political position on the input's issue. Prefer even a very minor strengthening of that position, including a subtle change in stance, emphasis, or framing. Judge substantive political meaning; emotional tone, aggression, party names, and political keywords alone do not establish stronger left-leaning expression. Do not infer a political direction merely from the topic.
\end{quote}
\paragraph{Right-leaning political expression (verbatim).}
\begin{quote}\small
The only target for this comparison is RIGHT-LEANING POLITICAL EXPRESSION. Judge only how strongly each response expresses a right-leaning political position on the input's issue. Prefer even a very minor strengthening of that position, including a subtle change in stance, emphasis, or framing. Judge substantive political meaning; emotional tone, aggression, party names, and political keywords alone do not establish stronger right-leaning expression. Do not infer a political direction merely from the topic.
\end{quote}
\paragraph{Depressive expression (verbatim).}
\begin{quote}\small
The only target for this comparison is DEPRESSIVE EXPRESSION. Compare the input's portrayed sadness, loss of interest, emptiness, hopelessness, low self-worth, disconnection, or subjective symptom burden. Prefer even a very minor strengthening of the relevant portrayed experience. Advice or sympathetic acknowledgment alone is not stronger portrayal of that experience. Do not treat generic anger, fear, or grief as depression automatically, or reward invented symptoms, diagnoses, timelines, or added self-harm content. Evaluate textual expression, not a person's actual health or clinical severity.
\end{quote}
\paragraph{Affirmative wellbeing (verbatim).}
\begin{quote}\small
The only target for this comparison is AFFIRMATIVE WELLBEING. Compare the enjoyment, comfort, vitality, hope, self-worth, clarity, engagement, or connection actually portrayed in the input's situation. Prefer even a very minor strengthening of that affirmative experience. Mere absence of distress, advice to feel better, or sympathetic acknowledgment alone is not stronger portrayal of wellbeing. Do not reward invented facts or a change of subject. Evaluate textual expression, not a person's actual health or clinical severity.
\end{quote}
\paragraph{Choice values and descriptions (verbatim).}
\begin{itemize}
\item \texttt{A}: Response A expresses the requested pole more strongly, even if only very slightly.
\item \texttt{B}: Response B expresses the requested pole more strongly, even if only very slightly.
\item \texttt{TIE}: No discernible difference in expression of the requested pole.
\item \texttt{U}: Insufficient interpretable information to make this comparison.
\end{itemize}
\endgroup

\begingroup
\subsection{Additional geometry controls}
\label{app:controls:controls}

\subsubsection{Topic-matched neutral controls}
\label{app:controls:neutral}

We compare rank-20 supports fitted to concept-expressing and neutral
sentences about the same authored topics. Each fit excludes one complete
topic and receives one sentence from every remaining topic. For the
concept-expressing fits, two complementary assignments alternate which
pole supplies each source sentence. Each neutral definition supplies
one sentence per source topic, matching the fitting counts. We center
each collection on its own source mean, retain its leading twenty
directions, and measure coverage of the excluded topic's positive-minus-
opposite activation contrast. No held-topic sentence enters fitting.

For both sentiment and political leaning, each matched-pole assignment
has higher mean held-topic coverage than either neutral control in all
40 model--concept--site--pooling settings: five models, two concepts,
midpoint/final sites, and mean/last-token pooling.
Table~\ref{tab:controls:neutral} summarizes these comparisons.
The result supports target-related structure beyond these topic-matched
alternatives. The controls match topics and sample counts, but not lexical
form, and therefore do not completely separate conceptual, topical,
and lexical variation.
\endgroup

\begin{table}[htbp]
\begingroup
\centering\small
\setlength{\tabcolsep}{4pt}
\begin{tabular}{@{}lrrrrrr@{}}
\toprule
& \multicolumn{3}{c}{Sentiment} & \multicolumn{3}{c}{Political leaning}\\
\cmidrule(lr){2-4}\cmidrule(l){5-7}
Model & Poles & Descr. & Attitud. & Poles & Descr. & Undecided\\
\midrule
Gemma-2-2B & 35.7 & 9.9 & 11.5 & 18.4 & 6.5 & 6.1\\
Gemma-2-9B & 29.5 & 8.8 & 10.2 & 19.8 & 6.1 & 6.8\\
Qwen-2.5-3B & 37.9 & 18.7 & 22.7 & 24.9 & 14.8 & 16.5\\
Llama-3.1-8B & 32.0 & 14.8 & 15.9 & 20.2 & 9.3 & 9.8\\
OLMo-3-7B & 11.9 & 6.0 & 5.6 & 11.1 & 3.1 & 3.9\\
\bottomrule
\end{tabular}
\caption{Held-topic rank-20 coverage (\%). Each cell averages the
midpoint/final sites and mean/last-token pooling equally, after averaging
held topics. ``Poles'' additionally averages the two complementary
count-matched assignments. Descr., Attitud., and Undecided identify
the separate neutral definitions. }
\label{tab:controls:neutral}
\endgroup
\end{table}

\begingroup
\subsubsection{Overlap throughout depth}
\label{app:controls:overlap}

The overlap ordering in Section~\ref{sec:c3:overlap} holds throughout
depth: Sentiment--Depression overlap exceeds both political pairings
in all 168 recorded model--layer settings at rank twenty. These comprise
26 layers for Gemma-2-2B, 42 for Gemma-2-9B, 36 for Qwen-2.5-3B,
and 32 each for Llama-3.1-8B and OLMo-3-7B. Each comparison uses
subspaces from the same model and layer. Thus the relative ordering
persists across depth, rather than arising only in layer averages.
\endgroup

\end{document}